\documentclass{llncs}

\usepackage[pdftex]{graphicx}
\graphicspath{{figs/}}
\usepackage{amsmath}
\usepackage{amssymb}
\usepackage{algorithm}
\usepackage{algpseudocode}
\usepackage{times}

\usepackage{enumerate}
\newcommand{\figref}[1]{Fig.\ref{figure:#1}}
\newcommand{\tabref}[1]{Table \ref{table:#1}}

\begin{document}

\title{
  Recognition of Heat-Induced Food State Changes\\
  by Time-Series Use of Vision-Language Model\\
  for Cooking Robot
} %% 調理ロボットのための視覚-言語モデルの時系列利用による食材の熱による状態変化の認識

\author{Naoaki Kanazawa, Kento Kawaharazuka, Yoshiki Obinata, \\ Kei Okada, and Masayuki Inaba}

\institute{The University of Tokyo, 7-3-1, Hongo, Bunkyo-ku, Tokyo, Japan, \\
\email{kanazawa@jsk.imi.i.u-tokyo.ac.jp}}

\maketitle              % typeset the title of the contribution

\begin{abstract}

Cooking tasks are characterized by large changes in the state of the food, which is one of the major challenges in robot execution of cooking tasks. In particular, cooking using a stove to apply heat to the foodstuff causes many special state changes that are not seen in other tasks, making it difficult to design a recognizer. In this study, we propose a unified method for recognizing changes in the cooking state of robots by using the vision-language model that can discriminate open-vocabulary objects in a time-series manner. We collected data on four typical state changes in cooking using a real robot and confirmed the effectiveness of the proposed method. We also compared the conditions and discussed the types of natural language prompts and the image regions that are suitable for recognizing the state changes.

\end{abstract}
\begin{figure}[h]
  \centering
  \includegraphics[width=1.0\columnwidth]{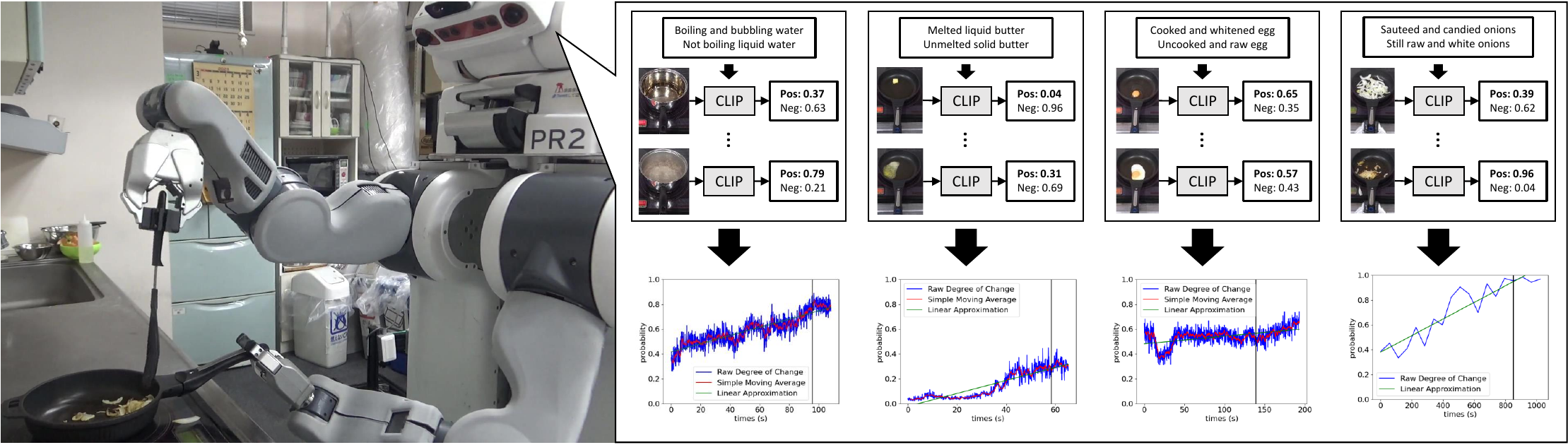}
  %% \vspace{-2mm}
  \caption{A cooking robot recognizes heat-induced food state changes. Using the vision-language model that can discriminate open vocabulary objects, the robot calculates the classification probability at each time by using two language descriptions that confirm or deny the change of state as prompts. The best prompt is selected by comparing the slope of a linear approximation of the positive classification probability, and the value smoothed by a simple moving average is used for threshold processing to recognize the state change.}
  %% 調理ロボットにより熱による食材の状態変化を認識する．オープンボキャブラリな物体識別を行うことが可能な視覚-言語モデルにより状態変化を肯定・否定する2つの言語記述をプロンプトとして各時刻の分類確率を計算する．肯定の分類確率の線形近似の傾きを比較して最良のプロンプトを選択し，単純移動平均により滑らかにした値を閾値処理することで状態変化を認識する．
  \label{figure:concept}
  %% \vspace{-6mm}
\end{figure}

\section{INTRODUCTION}

Cooking is one of the housework support tasks that robots are expected to perform. There are various issues regarding the execution of cooking tasks by robots in terms of action planning, manipulation, recognition, etc., and various approaches have been studied so far. \cite{beetz2011robotic,junge2020improving,kolathaya2018direct}.
In the cooking task, the foodstuff is the target object, and its state changes in a special way. This characteristic food ingredient state change is difficult to handle, and is one of the major challenges in the realization of cooking tasks by robots.
In particular, in cooking where heat is applied to food using a stove, the heat changes the state of the food physically and chemically, and there are various types of ingredients and state changes depending on the cooking recipe. Therefore, cooking robots that are used in the home need a recognition method that can handle the various state changes in a unified manner.

In cooking, it is necessary to recognize various state changes of ingredients, and it is difficult to prepare a large amount of data on all the state changes of all these ingredients. In this study, we use the vision-language model \cite{radford2021learning}, an open-vocablary object classification model, as a language-based image feature calculator to perform visual state change recognition based on linguistic descriptions of the state changes to be recognized.

In this paper, we propose a unified method for recognizing state changes from robot camera videos during cooking based on linguistic descriptions, by time-series use of the vision-language model. First, we classify the typical state changes of foods during heating and introduce a recognition method. Then, we verify the effectiveness of the proposed method through experiments using data acquired by a real robot, and discuss the results.

\section{RELATED WORK}

%% 加熱をする調理ロボット：
%% 食材に熱を加えて調理を行うロボットの研究としては，レシピ記述に基づいてパンケーキの調理を実行する研究\cite{beetz2011robotic}やバッチベイズ最適化によりオムレツ調理の品質最適化を行う研究\cite{junge2020improving}などが行われている．しかし，いずれの研究も熱による食材状態変化の認識は行っておらず，加熱時間により調整している．加熱時間による調整が必要な場合は有効な場合もあるが，同じ条件でも食材の個体差や実世界のランダム性により微妙な差異が存在するし，特に家庭で活躍する調理ロボットでは未知の料理レシピに対しても調理を行っていく必要があるため，食材の状態変化認識は重要である．

%% 食材の状態認識：
%% 食材の状態を認識する研究としては，専用に作成したデータセットによりCNNの学習を行うことで主に食材の切断状態について認識する研究\cite{paul2018classifying, jelodar2018identifying, sakib2021cooking}が行われてきた．しかし，調理ではレシピによって認識したい無数の状態や状態変化が存在するため，それら全ての食材の全ての状態変化についてのデータを大量に準備するのは困難である．また，これらの研究は調理工程終了後の状態を分類する問題を想定していて，リアルタイムな状態変化の変化点の認識の問題にはなっていない．

\subsubsection{Robot Cooking with Heat.}
Robots that cook food with heat have been researched, such as the pancake cooking robot based on recipe descriptions \cite{beetz2011robotic} and the omelette cooking quality optimization based on batch Bayesian optimization \cite{junge2020improving}.
However, none of these studies recognized food state changes caused by heat, and instead made adjustments based on the heating time.
Although adjustment by heating time may be necessary or effective in some cases, it is important to recognize changes in the ingredients' state because even under the same conditions, there are subtle differences due to the individual differences of ingredients and the randomness of the real world, and because cooking robots that work at home need to cook even for unknown recipes.

\vspace{-6mm}
\subsubsection{Recognition of Food State.}
Several studies have been conducted to recognize the state of food ingredients \cite{paul2018classifying,jelodar2018identifying,sakib2021cooking,takata2022recipe}, mainly in terms of the cutting state of the ingredients, by training CNNs on a specially created dataset. However, it is difficult to prepare a large amount of data for all the state changes of all these ingredients, because there are countless state changes that need to be recognized depending on the recipe. In addition, these studies assume the problem of classifying the states after the cooking process is completed, and do not address the problem of recognizing the change points of real-time state changes during cooking.
A similar problem setting can be found in capturing human cooking videos \cite{shi2019dense,huang2020multimodal,nishimura2021state}, but the problem is reversed because the cooking robot wants to recognize that a state change specified in a recipe has occurred.

\vspace{-6mm}
%% 似た問題を扱った研究例として，人の調理動画のcaptioningが挙げられるが，調理ロボットではレシピテキストで与えられた状態変化が発生したことを認識したいので逆の問題設定になっている．

\subsubsection{Vision-Language Model.}
%% 近年，インターネット上の膨大なテキスト関連データを用いた大規模事前学習済みモデルがさかんに開発されている．中でも，視覚-言語モデルは言語記述に基づいてオープンボキャブラリーな画像認識タスクを行うことができる．画像分類やセマンティックセグメンテーション，物体検出を行うことができるモデルや複数の画像認識タスクを統一的に解くことが可能なフレームワークなどが提案されている．我々はこの視覚-言語モデルにいち早く着目しロボットビジョン及びロボットプログラミングの新しい可能性を模索している．

In recent years, many large-scale pre-trained models have been developed using the vast amount of text-related data available on the Internet. Among them, vision-language models can perform open-vocabulary image recognition tasks based on linguistic descriptions. Models that can perform image classification \cite{radford2021learning}, semantic segmentation \cite{li2022lseg}, and object detection \cite{zhou2022detecting} have been proposed, as well as frameworks that can solve multiple image recognition tasks in a unified manner \cite{wang2022ofa}. We are exploring new possibilities for robot vision and robot programming by focusing on these vision-language models \cite{kawaharazuka2023vlm}.

\section{COOKING STATE CHANGE RECOGNITION BY TIME-SERIES USE OF VISION-LANGUAGE MODEL}

\subsection{Classification of Heat-Induced Food State Changes}

%% 加熱調理では鍋やフライパンなどに食材を乗せて加熱することで食材を状態変化させる．ロボットが加熱調理を実行できるようにするには，その状態変化を認識する機能を持つことが重要である．加熱によって生じる食材の状態変化は物理的変化と化学的変化の2つが存在する．物理的変化は相転移のことである．加熱により発生する相転移は気化・融解・昇華の3つである．昇華はフリーズドライを作成する際に真空状態で昇華乾燥するなどの場面であり家庭での調理では一般的でないため今回は対象外とし，気化と融解の2つの物理的変化について扱う．調理における化学的変化としては，おもにタンパク質の熱変性と焼き色がつくなどのメイラード反応の2つがある．4つの状態変化それぞれの特徴については以下の通りである．
In cooking, food is placed on a pot or pan and heated to change its state. In order for robots to be able to perform cooking, it is important that they have the ability to recognize this state change. There are two types of heat induced foodstuff state changes: physical and chemical. The physical change is the phase transition. There are three types of phase transitions that occur with heating: vaporization, melting, and sublimation. Sublimation is not a common process in home cooking, and is not included in this study, but rather two physical changes, vaporization and melting, are discussed. The main chemical changes that occur in cooking are thermal denaturation of proteins and Maillard reactions such as browning. The characteristics of each of the four state changes are described below.

\vspace{-6mm}
\subsubsection{Vaporization.}
%% 気化では液体である対象物が加熱されることで沸点に達し，気体に変化する．加熱調理における気化の代表例としては，水を鍋に入れて沸騰させる調理などが挙げられる．水が沸騰する際には，水蒸気が激しく発生し水の中から泡が出てくるのが特徴的である．
In vaporization, a liquid object is heated to its boiling point and transformed into a gas. A typical example of vaporization in cooking is when water is placed in a pot and brought to a boil. When water boils, it is characteristic that vapor is generated violently and bubbles emerge from the water.

\vspace{-6mm}
\subsubsection{Melting.}
%% 融解では固体である対象物が加熱されることで融点に達し，液体に変化する．加熱調理における融解の代表例としては，バターをフライパンで熱して溶かすという調理工程が挙げられる．バターが溶ける際には，固体の塊だったバターが次第に溶けていき全体が液体になるという変化が発生する．
In melting, an object that is solid is heated to its melting point and transformed into a liquid. A typical example of melting in cooking is the process of heating butter in a frying pan to melt it. When butter melts, the solid mass of butter gradually melts and becomes entirely liquid.

\vspace{-6mm}
\subsubsection{Thermal Denaturation of Proteins.}
%% タンパク質の熱変性では，タンパク質を構成するアミノ酸の立体的な構造が熱により破壊され，性質が変化する．加熱調理におけるタンパク質の熱変性としては卵の凝固が挙げられる．卵は変性温度に達すると固まり，特に卵白は白く変化する．
In thermal denaturation of proteins, the three-dimensional structure of the amino acids that make up the protein is destroyed by heat, and the properties of the protein change. One example of thermal denaturation of proteins in cooking is the coagulation of eggs. When eggs reach the denaturation temperature, they solidify, and egg whites in particular turn white.

\vspace{-6mm}
\subsubsection{Maillard Reaction.}
%% メイラード反応では，カルボニル化合物とアミノ化合物が熱により反応し，褐色物質のメラノイジンや香気成分が生成される．加熱調理におけるメイラード反応の代表例としては，玉ねぎを飴色になるまで炒める調理が挙げられる．
In the Maillard reaction, carbonyl compounds and amino compounds react with heat to produce the brown substance melanoidin and aroma components. A typical example of the Maillard reaction in cooking is frying onions until they become candied.

\subsection{Method for Designing State Recognizer for Cooking Using Vision-Language Model}
%% 視覚-言語モデルを使った加熱調理の状態認識器の設計法

%% 本研究では，オープンボキャブラリーな物体識別モデルである視覚-言語モデルCLIPを利用して食材の状態変化認識法を提案する．状態変化を肯定・否定する2つの言語記述をプロンプトとして各時刻における画像の分類確率を計算し，肯定の分類確率を状態変化度として時系列データ処理を行うことで状態認識を行う．その際には，視覚-言語モデルのプロンプトとしてどのような言語記述を選択するか，と時系列データ処理の方法が重要になる．

%% 本研究では，まず対象とする食材の状態変化の言語記述として以下の4種類の言語記述を考えた．
%% (a)状態変化をシンプルに記述した食材単語で終わる形の言語記述
%% (b)状態変化により起こる変化に関する記述も追加した食材単語で終わる形の言語記述
%% (c)状態変化をシンプルに記述した食材単語で始まる形の言語記述
%% (d)状態変化により起こる変化に関する記述も追加した食材単語で始まる形の言語記述
%% それぞれの言語記述について，肯定と否定の2つの言語記述を用意しプロンプトとして利用する．対象となる状態変化のデータに対してこの4つのプロンプトを用いて状態変化度を前述の方法で計算し，時系列データ分析を行うことで最良のプロンプトを選択する．データ分析では，計算された状態変化度の時系列データを線形近似し，線形近似した直線の傾きを比較する．その傾きが最も大きくなるプロンプトを認識器設計に適したプロンプトであると評価する．

%% また，選択されたプロンプトに対する認識器としての性能評価を行うために，単純な時系列処理による状態変化認識法についても提案する．状態変化の動画データに対して，人が変化が発生したと感じた時刻を記録し，その時刻における時系列データの単純移動平均値を閾値として設定し，未知の状態変化の動画データに対して単純移動平均値に対する閾値処理により状態変化認識を行う．

In this study, we propose a unified method for recognizing food ingredients' state change using the vision-language model CLIP \cite{radford2021learning}, which is an open vocabulary object classification model. The model calculates the classification probability of an image at each time using two linguistic descriptions that confirm or deny the state change as prompts, and performs state recognition by time series data processing using the positive classification probability. In this process, it is important what kind of linguistic descriptions are selected as prompts for the vison-language model and how the time-series data processing is performed.

We considered the following four types of linguistic descriptions of target foodstuff state changes.
%% \begin{enumerate}[(a)]
%%  \item Language descriptions in the form of ingredient word ending with simple description of the state change, e.g. \textbf{Boiling water}.
%%  \item Language descriptions in the form of ingredient word ending with additional description of changes caused by the state change, e.g. \textbf{Boiling and bubbling water}.
%%  \item Language descriptions beginning with the ingredient word that simply describes the state change, e.g. \textbf{Water that is boiling}.
%%  \item Language descriptions beginning with the ingredient word that include additional description of changes caused by the state change, e.g. \textbf{Water that is boiling and bubbling}.
%% \end{enumerate}
\begin{enumerate}[(a)]
 \item Simple description of the state change, e.g. \textbf{Boiling water}.
 \item Language descriptions with additional description of changes caused by the state change, e.g. \textbf{Boiling and bubbling water}.
 \item Language descriptions with the ingredient word at the beginning that simply describes the state change, e.g. \textbf{Water that is boiling}.
 \item Language descriptions with the ingredient word at the beginning that include additional description of changes caused by the state change, e.g. \textbf{Water that is boiling and bubbling}.
\end{enumerate}
Hypothetically, prompts with detailed descriptions of changes are more sensitive to changes than simple descriptions, making them more suitable as recognizers. For each type, two language descriptions (positive and negative) are prepared and used as prompts. The degree of state change is calculated using these four prompts for the target state change data in the manner described above, and the best prompt is selected through time series data analysis. In the data analysis, the time-series data of the calculated degree of state change is linearly approximated, and the slopes of approximated lines (LA Slope) are compared. The prompt with the largest slope is evaluated as a suitable prompt for the recognizer design.

In order to evaluate the performance as a recognizer, we also propose a state change recognition method based on simple time series processing. For each video data of the state change, we record the time at which a person perceives the change to have occurred.
Simple moving average with window size 10 
of the time-series data at that time is set as a threshold value, and state change recognition is performed on unknown video data by threshold processing.
%% Simple moving averages are performed with a window size of 10 for 10hz data. %% 追加 この際の単純移動平均は10hzのデータに対してウィンドウサイズを10として行っている．

%% \begin{figure}[thpb]
%%   \centering
%%   \includegraphics[width=0.7\columnwidth]{figs/ias18-cook-state-rec-system}
%%   \caption{system}
%%   \label{figure:system}
%% \end{figure}

\section{EXPERIMENTS}

%% \subsection{Heat-Induced State Change Recognition Experiment}
%% 台車移動型ロボットPR2のカメラで代表的な4つの加熱調理中の食材状態変化の動画を記録し，そのデータを用いて提案手法の有効性を検証した．画像の注視領域も重要であると考えられるため，鍋やフライパン全体を囲む矩形領域と鍋やフライパンの中身のみを囲む矩形領域の2種類の範囲を切り抜いて利用した．状態変化について，4種類のプロンプトの比較を行い，選択された最良プロンプトを用いて既知データと同じ火力のデータと異なる火力のデータの2条件に対して状態変化認識器としての性能評価も行った．

We recorded videos of food state changes during four typical heating cooking processes using the camera of the cart-mobile robot PR2, and used the data to verify the effectiveness of the proposed method. Since the gazing area of the image is also considered important, we used two types of cropped areas: a rectangular area surrounding the entire pot or pan, and a rectangular area surrounding only the contents of the pot or pan. It is thought that it is easier to recognize the state change if one gazes only at the contents, because the object in which the state change occurs can be centrally imaged. We compared four types of prompts for the state change, and evaluated the performance of the state change recognizer using the selected best prompt for two conditions: data with the same heat power as the known data and data with different heat power.

\subsubsection{Vaporization.}
%% 気化の状態変化認識実験として，鍋に水を入れて沸騰させる調理のデータを記録し提案手法の評価を行った．
%% 記録した状態変化データの1つの画像列と人が状態変化が発生したと感じた時刻の画像を\figref{images_vaporization}に示す．
%% 水の沸騰の認識プロンプトとして4種類のプロンプトを用意し，そのデータに対して視覚-言語モデルによる状態変化度の計算と時系列データ分析により比較を行った．(\figref{graph_vaporization}, \tabref{comparison_vaporization})

\begin{figure}[htb]
  \begin{center}
    \begin{minipage}{0.1\columnwidth}
      \includegraphics[height=1.6cm]{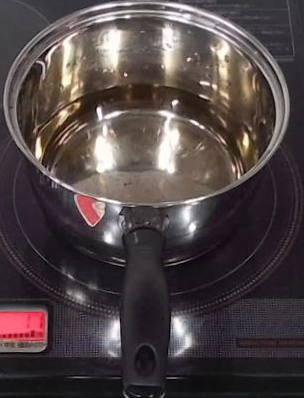}
      \centering 0s
    \end{minipage}
    %% \hspace{0.01\columnwidth}
    \begin{minipage}{0.1\columnwidth}
      \includegraphics[height=1.6cm]{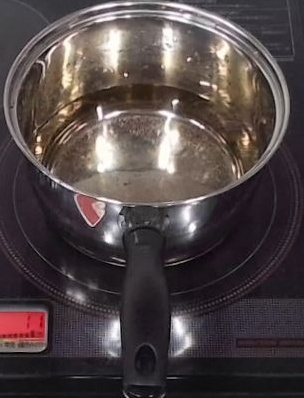}
      \centering 20s
    \end{minipage}
    %% \hspace{0.01\columnwidth}
    \begin{minipage}{0.1\columnwidth}
      \includegraphics[height=1.6cm]{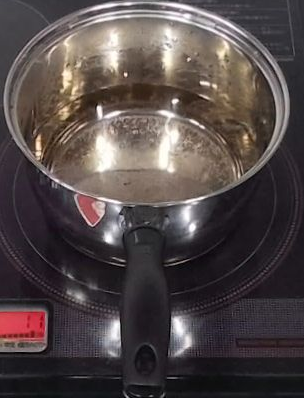}
      \centering 40s
    \end{minipage}
    %% \hspace{0.01\columnwidth}
    \begin{minipage}{0.1\columnwidth}
      \includegraphics[height=1.6cm]{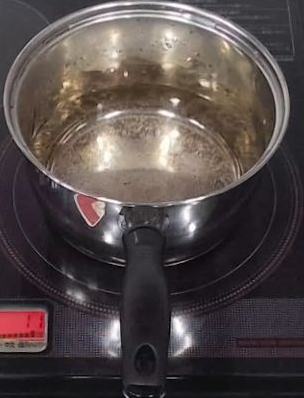}
      \centering 60s
    \end{minipage}
    %% \hspace{0.01\columnwidth}
    \begin{minipage}{0.1\columnwidth}
      \includegraphics[height=1.6cm]{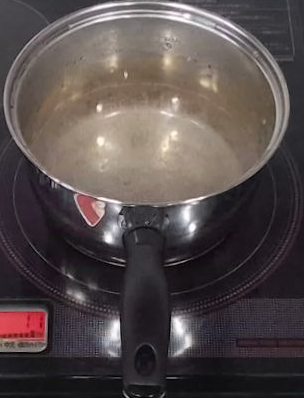}
      \centering 80s
    \end{minipage}
    %% \hspace{0.01\columnwidth}
    \begin{minipage}{0.1\columnwidth}
      \includegraphics[height=1.6cm]{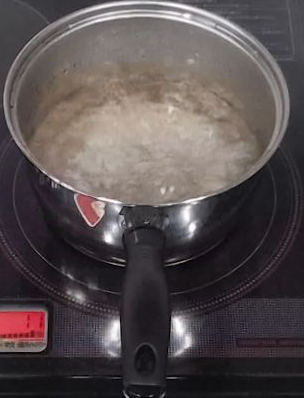}
      \centering 100s
    \end{minipage}
    \hspace{0.01\columnwidth}
    \begin{minipage}{0.1\columnwidth}
      \includegraphics[height=1.6cm]{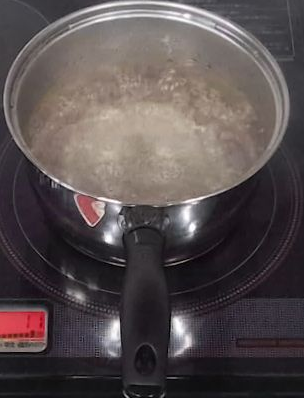}
      \centering (A)
    \end{minipage}
    %% \hspace{0.01\columnwidth}
    \begin{minipage}{0.17\columnwidth}
      \includegraphics[height=1.6cm]{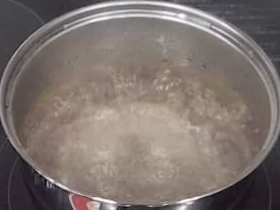}
      \centering (B)
    \end{minipage}
    %% \hspace{0.01\columnwidth}
  \end{center}
  \vspace{-6mm}
  \caption{State change of vaporization. (A) Image of the entire pot at the time when the person felt the state change occurred (95.7s). (B) Image of only the contents of the pot at the same time.}
  %% 水の沸騰の状態変化の様子．(A)人が状態変化が発生したと感じた時刻(95.7s)の鍋全体の画像．(B)同じ時刻の鍋の中身のみの画像
  \vspace{-12mm}
  \label{figure:images_vaporization}
\end{figure}

\begin{figure}[h!]
  \begin{center}
    \begin{minipage}{0.24\columnwidth}
      \includegraphics[width=\columnwidth]{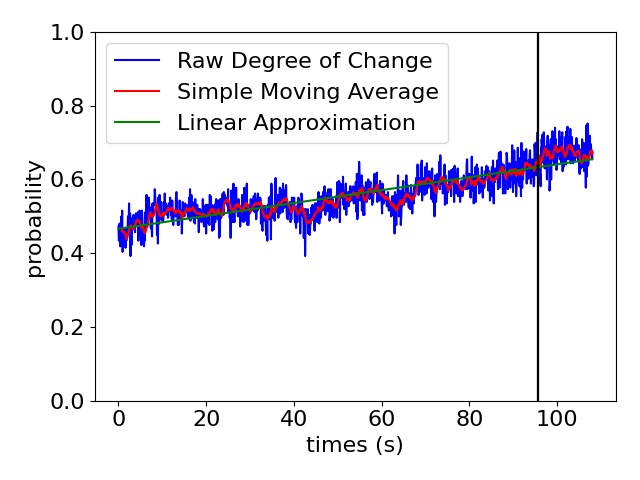}
      \centering (a)-entire
    \end{minipage}
    %% \hspace{0.01\columnwidth}
    \begin{minipage}{0.24\columnwidth}
      \includegraphics[width=\columnwidth]{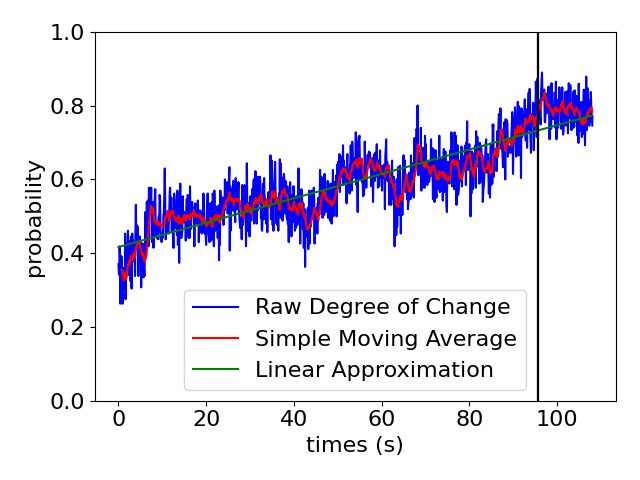}
      \centering (b)-entire
    \end{minipage}
    %% \hspace{0.01\columnwidth}
    \begin{minipage}{0.24\columnwidth}
      \includegraphics[width=\columnwidth]{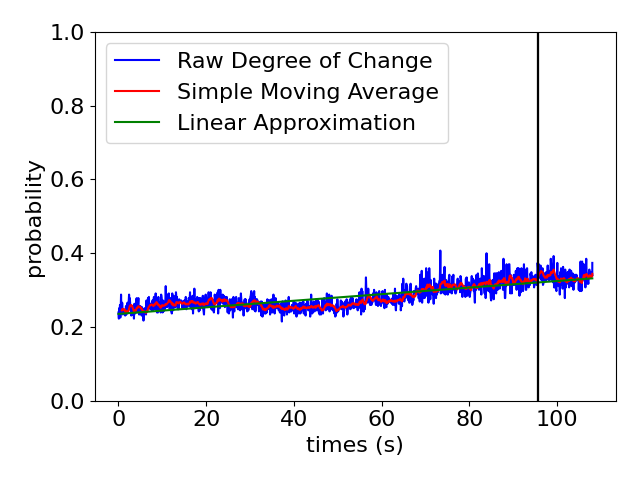}
      \centering (c)-entire
    \end{minipage}
    %% \hspace{0.01\columnwidth}
    \begin{minipage}{0.24\columnwidth}
      \includegraphics[width=\columnwidth]{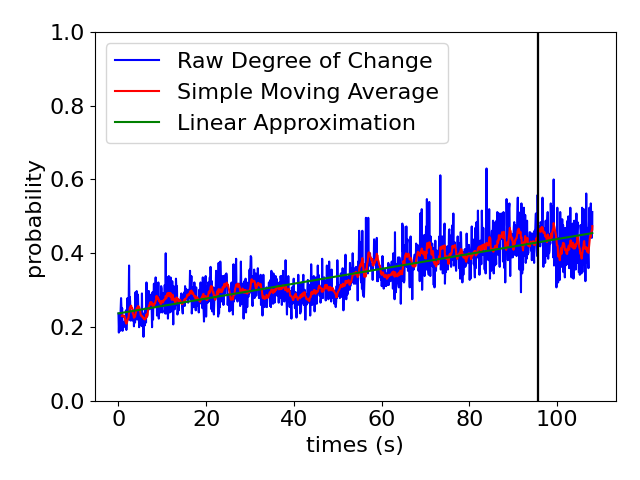}
      \centering (d)-entire
    \end{minipage}
    %% \hspace{0.01\columnwidth}
    \begin{minipage}{0.24\columnwidth}
      \includegraphics[width=\columnwidth]{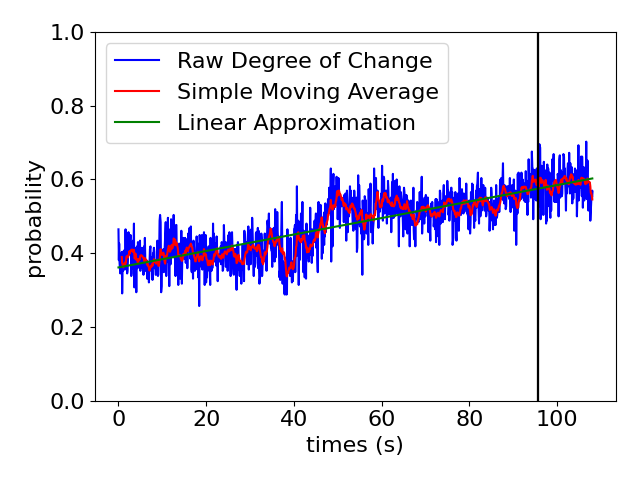}
      \centering (a)-contents
    \end{minipage}
    %% \hspace{0.01\columnwidth}
    \begin{minipage}{0.24\columnwidth}
      \includegraphics[width=\columnwidth]{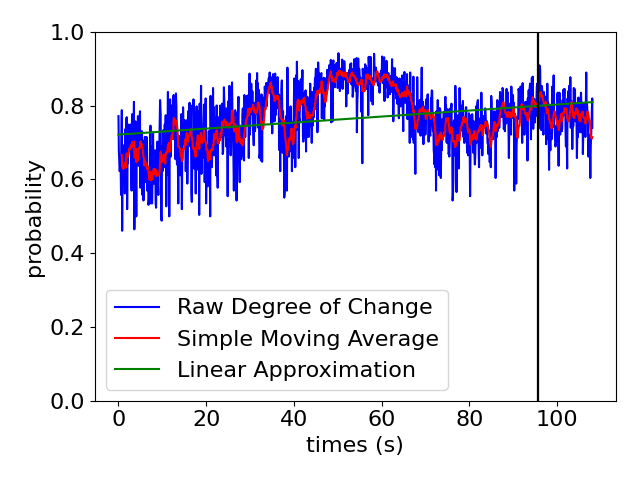}
      \centering (b)-contents
    \end{minipage}
    %% \hspace{0.01\columnwidth}
    \begin{minipage}{0.24\columnwidth}
      \includegraphics[width=\columnwidth]{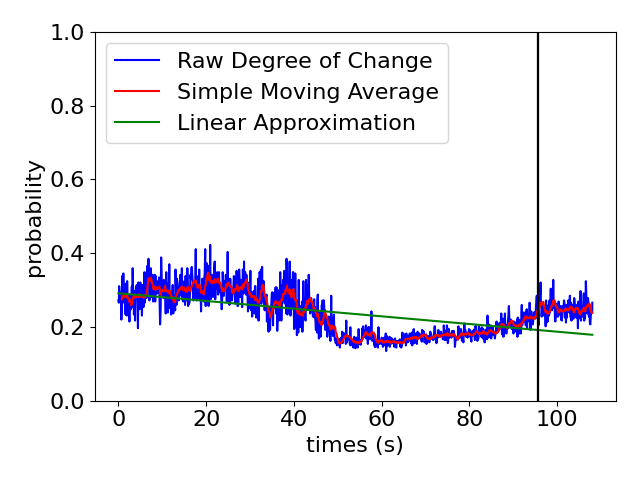}
      \centering (c)-contents
    \end{minipage}
    %% \hspace{0.01\columnwidth}
    \begin{minipage}{0.24\columnwidth}
      \includegraphics[width=\columnwidth]{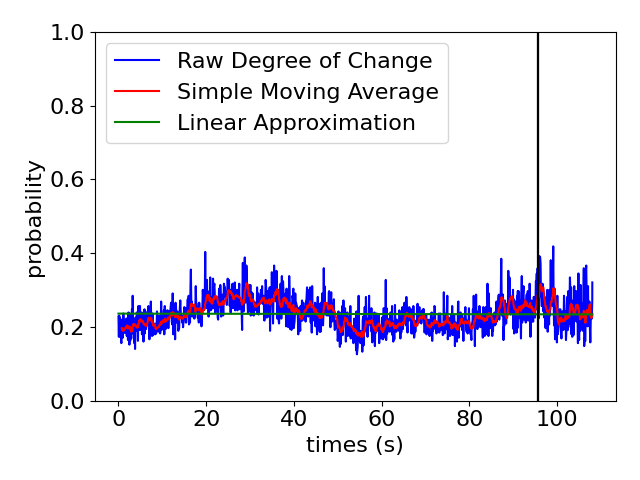}
      \centering (d)-contents
    \end{minipage}
    %% \hspace{0.01\columnwidth}
  \end{center}
  \vspace{-6mm}
  \caption{Plots of inferred degree of state change for each condition against vaporization data.}
  %% 気化データに対するそれぞれの条件での状態変化度の推論結果のプロット
  \vspace{-6mm}
  \label{figure:graph_vaporization}
\end{figure}

We evaluated the proposed method by recording data from a cooking session in which water was placed in a pot and brought to a boil as a vaporization state change recognition experiment.
One image sequence of the recorded data and the image at the time when the person felt the state change occurred are shown in \figref{images_vaporization}.
Four types of prompts were prepared as recognition prompts for boiling water, and a comparison was made by calculating the degree of state change using the vison-language model and time-series data analysis for the data. (\tabref{comparison_vaporization}, \figref{graph_vaporization})

\begin{table}[h]
  \begin{center}
    \caption{Comparison of prompts and gazing areas for vaporization data.}
    \vspace{-3mm}
    \label{table:comparison_vaporization}
    \scalebox{0.8}{
      \begin{tabular}{|c|c|c|c||c|}
        \hline
        & Gaze Area & Positive Prompt & Negative Prompt & LA Slope  \\
        \hline
        (a)-entire & entire pot & Boiling water & Not boiling water & 0.00175 \\
        (b)-entire & entire pot & Boiling and bubbling water & Not boiling liquid water & 0.00331 \\
        (c)-entire & entire pot & Water that is boiling & Water that is not boiling & 0.00089 \\
        (d)-entire & entire pot & Water that is boiling and bubbling & Water that is not boiling and liquid & 0.00202 \\
        \hline
        (a)-contents & contents & Boiling water & Not boiling water & 0.00233 \\
        (b)-contents & contents & Boiling and bubbling water & Not boiling liquid water & 0.00082 \\
        (c)-contents & contents & Water that is boiling & Water that is not boiling & -0.00104 \\
        (d)-contents & contents & Water that is boiling and bubbling & Water that is not boiling and liquid & -0.00002 \\
        \hline
    \end{tabular}}
    \vspace{-16mm}
  \end{center}
\end{table}

%% 比較の結果，鍋全体を注視領域とする場合においては(b)のプロンプトが最良であり，鍋の中身のみを注視領域とする場合においては(a)のプロンプトが最良であった．それぞれの注視条件において，最良であると選択されたプロンプトと閾値を用いて未知データに対する状態変化判定を行い，認識器としての性能を評価した．認識器を設計したのと同じ火力のデータと異なる火力のデータに対して状態変化認識を行い，認識器が判定した状態変化時刻と人のアノテーション時刻との差分を比較した (\tabref{result})．気化のデータにおいてはどちらの火力条件のデータについても鍋全体を注視領域とした方が認識器の性能がよくなるという結果になった．

\begin{table}[h!]
  \begin{center}
    \caption{State change recognition results for unknown data of vaporization.}
    \label{table:result_vaporization}
    \vspace{-3mm}
    \scalebox{0.9}{
      \begin{tabular}{|c||c|c|}
        \hline
        & Same Power Diff (s) &  Different Power Diff (s)  \\
        \hline
        (b)-entire & 1.6 & 6.5 \\
        \hline
        (a)-contents & 51.4 & 48.9 \\
        \hline
    \end{tabular}}
    \vspace{-6mm}
  \end{center}
\end{table}

\begin{figure}[h!]
  \begin{center}
    \begin{minipage}{0.48\columnwidth}
      \includegraphics[width=\columnwidth]{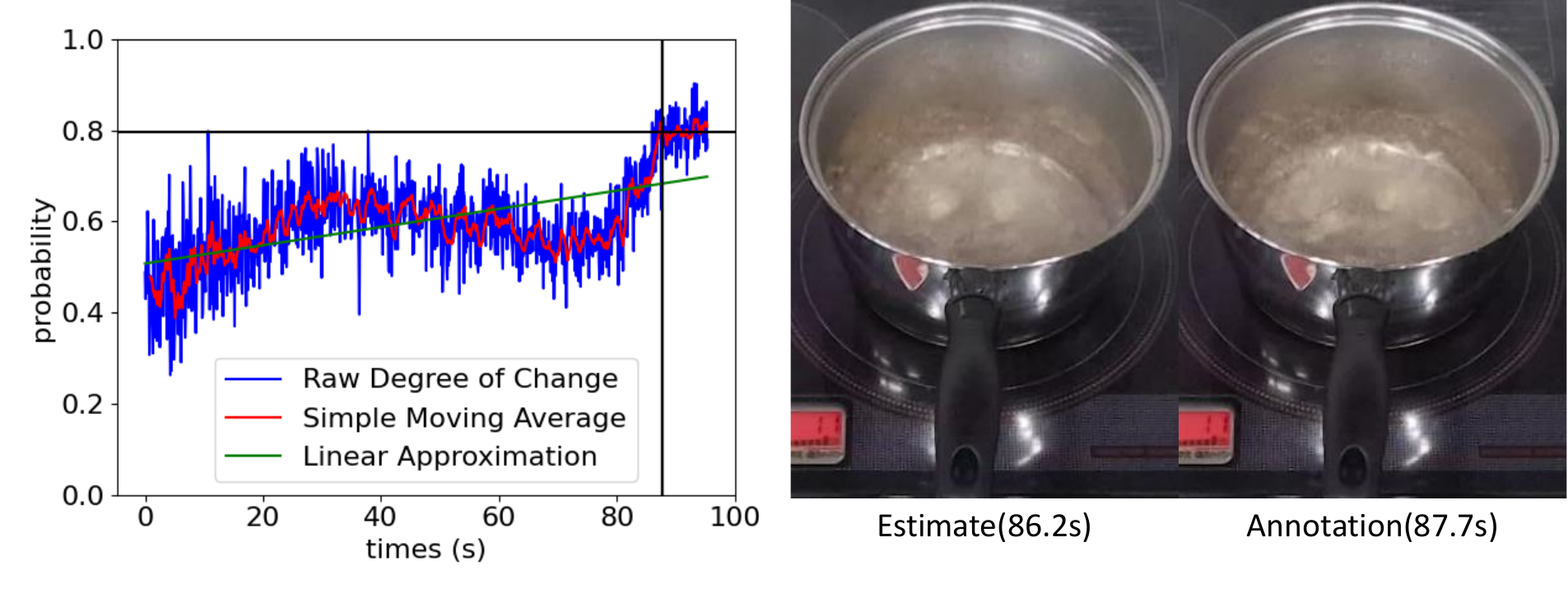}
      \centering same-power-(b)-entire
    \end{minipage}
    %% \hspace{0.01\columnwidth}
    \begin{minipage}{0.48\columnwidth}
      \includegraphics[width=\columnwidth]{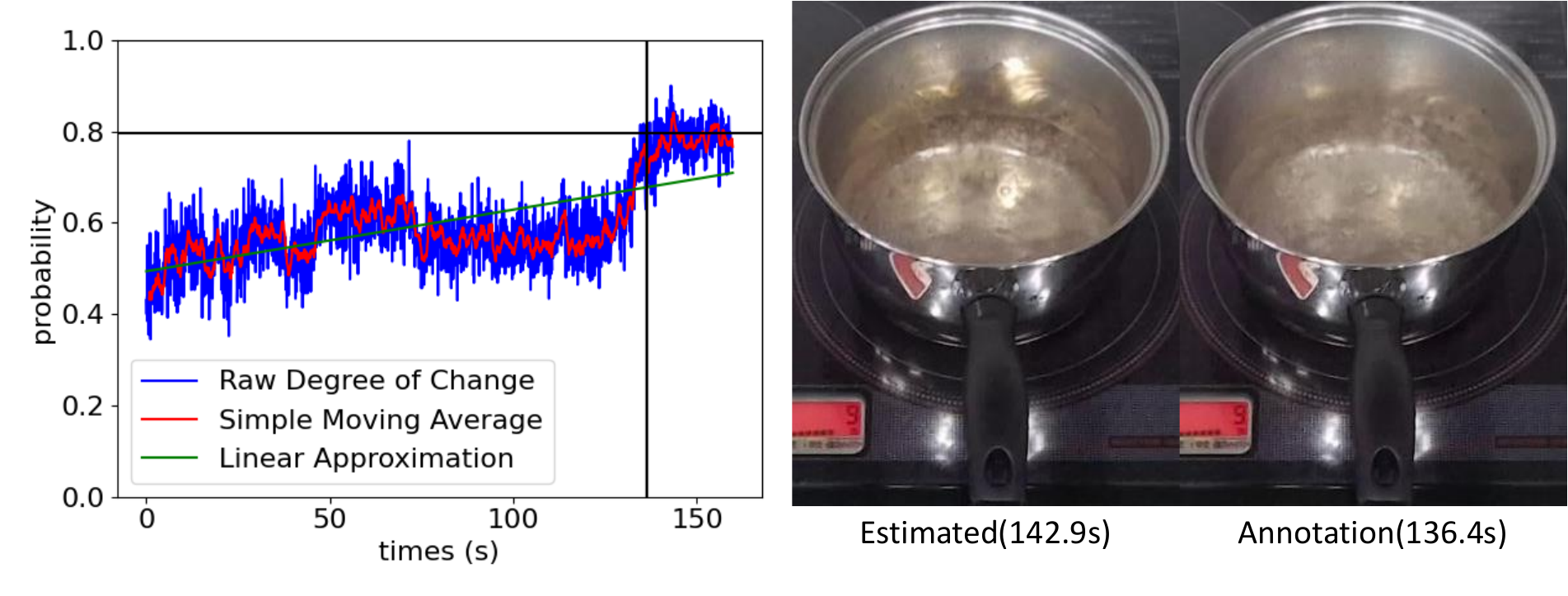}
      \centering different-power-(b)-entire
    \end{minipage}
    %% \hspace{0.01\columnwidth}
    \begin{minipage}{0.48\columnwidth}
      \includegraphics[width=\columnwidth]{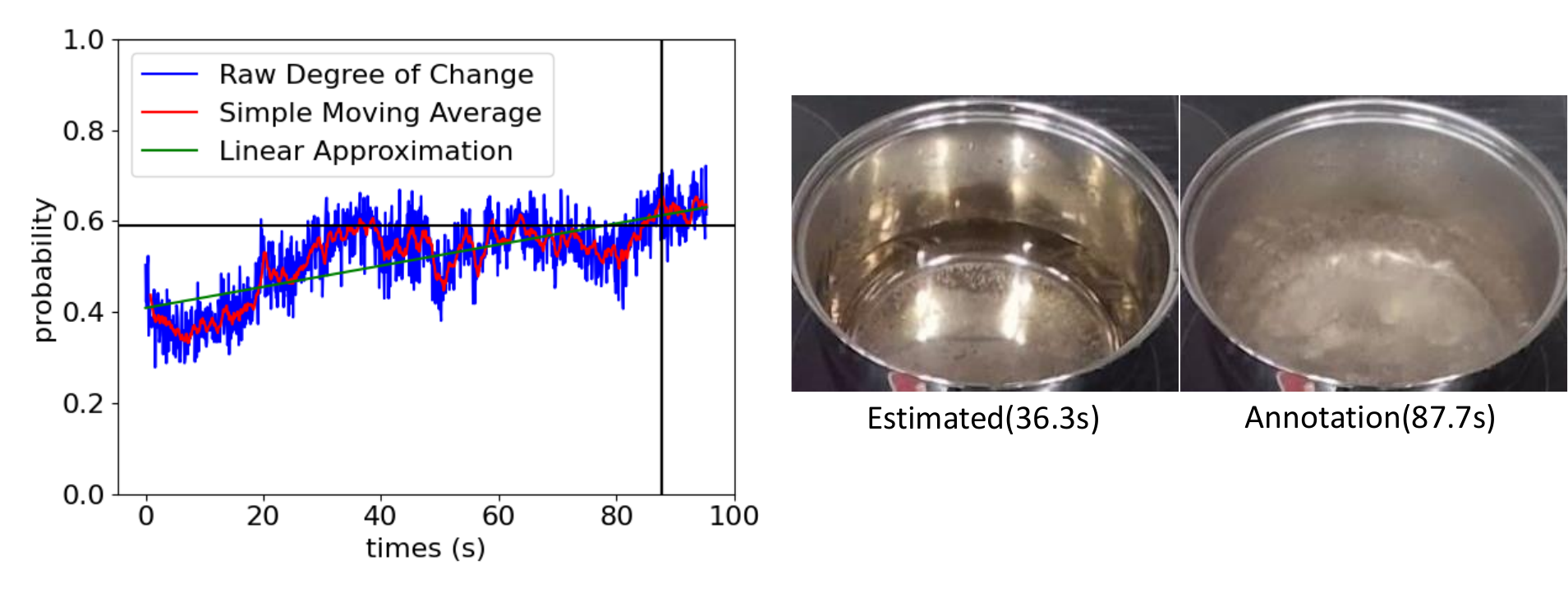}
      \centering same-power-(a)-contents
    \end{minipage}
    %% \hspace{0.01\columnwidth}
    \begin{minipage}{0.48\columnwidth}
      \includegraphics[width=\columnwidth]{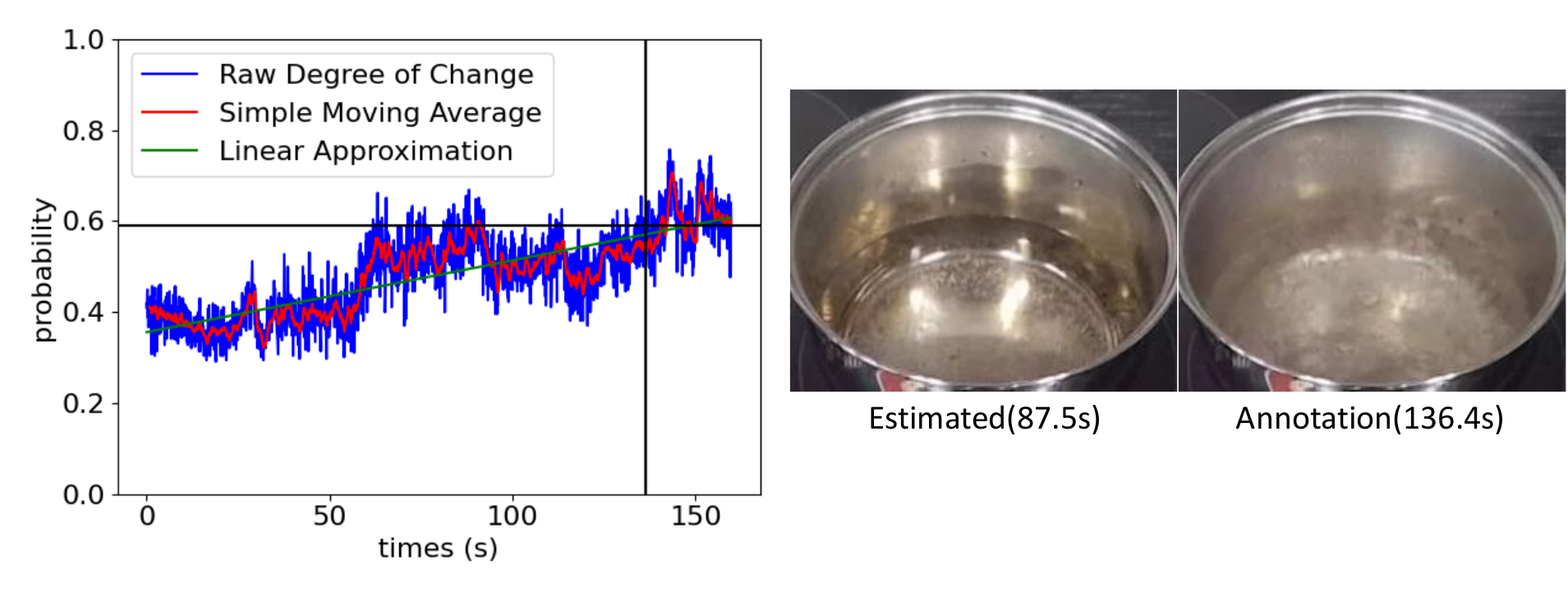}
      \centering different-power-(a)-contents
    \end{minipage}
    %% \hspace{0.01\columnwidth}
  \end{center}
  \vspace{-6mm}
  \caption{Plots of state change recognition results for unknown vaporization data and comparison of images at the estimated time and at the time annotated by human.}
  %% 未知の気化データに対する状態変化認識結果のグラフと推定された時刻の画像と人がアノテーションした時刻の画像の比較
  \vspace{-6mm}
  \label{figure:result_image_vaporization}
\end{figure}

As a result of the comparison, prompt (b) was the best when the entire pot was the gazing area, and prompt (a) was the best when only the contents of the pot were the gazing area. In each gazing condition, the prompt selected as the best and the threshold values were used to make state change judgments for unknown data, to evaluate the performance of the recognizer. State change recognition was performed on data of the same and different firepower for which the recognizer was designed, and the difference between the state change time determined by the recognizer and the annotation time of the person was compared (\tabref{result_vaporization}, \figref{result_image_vaporization}). In the case of the vaporization, the performance of the recognizer was better when the entire pot was used as the gazing area for the data of both thermal power conditions.

\subsubsection{Melting.}
%% 融解の状態変化については，フライパンでバターを加熱し溶かす調理を行いデータを収集した(\figref{})．融解についても同様に4種類のプロンプトを設計し，それぞれの条件での比較を行った．バターが溶けるデータについては，どちらの注視条件でも(b)のプロンプトが最良であるという結果になった．最良である(b)のプロンプトを使った認識器性能評価の結果を\tabre{result}に示す．融解のデータについては，同じ火力の場合は中身のみを注視した方がよく，違う火力のデータについては鍋全体を見た方が良いという結果になった．

\begin{figure}[h!]
  \begin{center}
    \begin{minipage}{0.09\columnwidth}
      \includegraphics[height=1.6cm]{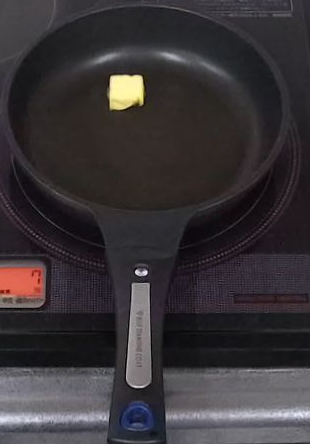}
      \centering 0s
    \end{minipage}
    %% \hspace{0.01\columnwidth}
    \begin{minipage}{0.09\columnwidth}
      \includegraphics[height=1.6cm]{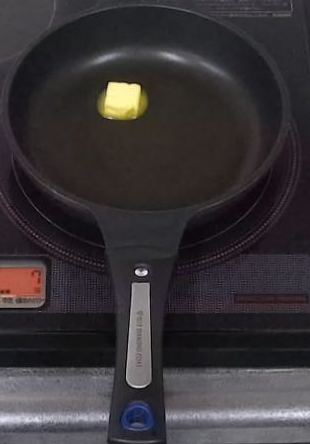}
      \centering 10s
    \end{minipage}
    %% \hspace{0.01\columnwidth}
    \begin{minipage}{0.09\columnwidth}
      \includegraphics[height=1.6cm]{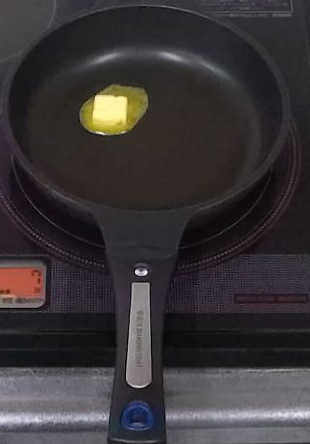}
      \centering 20s
    \end{minipage}
    %% \hspace{0.01\columnwidth}
    \begin{minipage}{0.09\columnwidth}
      \includegraphics[height=1.6cm]{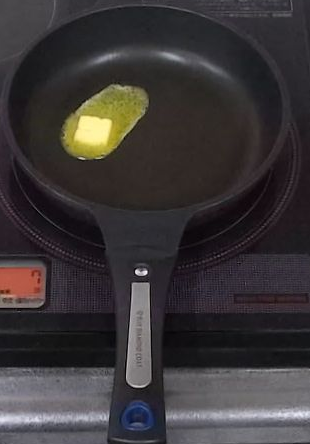}
      \centering 30s
    \end{minipage}
    %% \hspace{0.01\columnwidth}
    \begin{minipage}{0.09\columnwidth}
      \includegraphics[height=1.6cm]{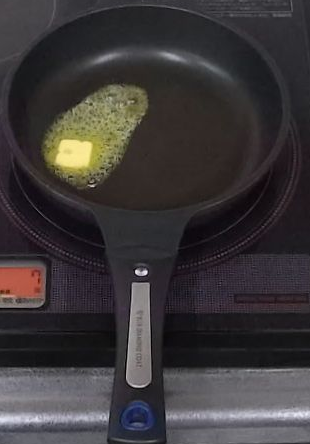}
      \centering 40s
    \end{minipage}
    %% \hspace{0.01\columnwidth}
    \begin{minipage}{0.09\columnwidth}
      \includegraphics[height=1.6cm]{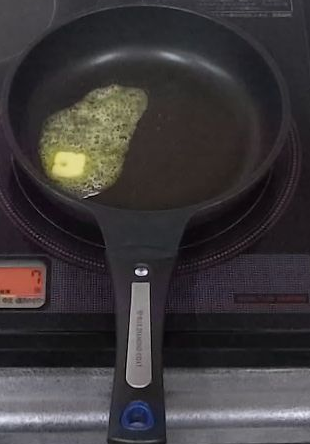}
      \centering 50s
    \end{minipage}
    %% \hspace{0.01\columnwidth}
    \begin{minipage}{0.09\columnwidth}
      \includegraphics[height=1.6cm]{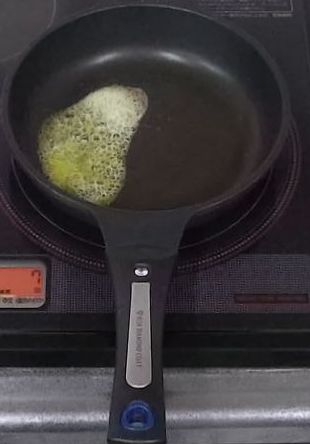}
      \centering 60s
    \end{minipage}
    \hspace{0.01\columnwidth}
    \begin{minipage}{0.09\columnwidth}
      \includegraphics[height=1.6cm]{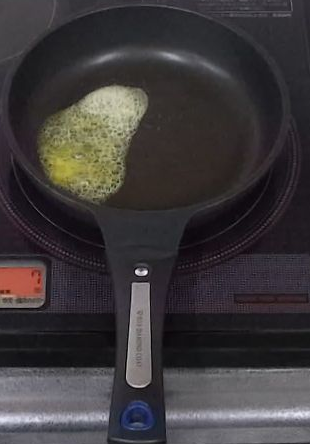}
      \centering (A)
    \end{minipage}
    %% \hspace{0.01\columnwidth}
    \begin{minipage}{0.16\columnwidth}
      \includegraphics[height=1.6cm]{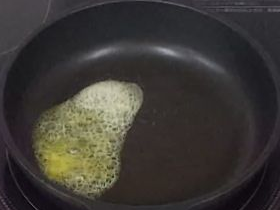}
      \centering (B)
    \end{minipage}
    %% \hspace{0.01\columnwidth}
  \end{center}
  \vspace{-6mm}
  \caption{State change of melting. (A) Image of the entire pot at the time when the person felt the state change occurred (58.3s). (B) Image of only the contents of the pot at the same time.}
  \label{figure:images_melting}
  \vspace{-2mm}
\end{figure}

\begin{figure}[h!]
  \begin{center}
    \begin{minipage}{0.24\columnwidth}
      \includegraphics[width=\columnwidth]{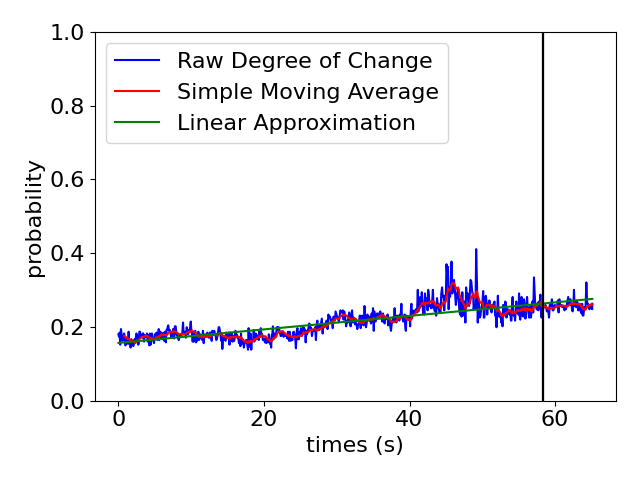}
      \centering (a)-entire
    \end{minipage}
    %% \hspace{0.01\columnwidth}
    \begin{minipage}{0.24\columnwidth}
      \includegraphics[width=\columnwidth]{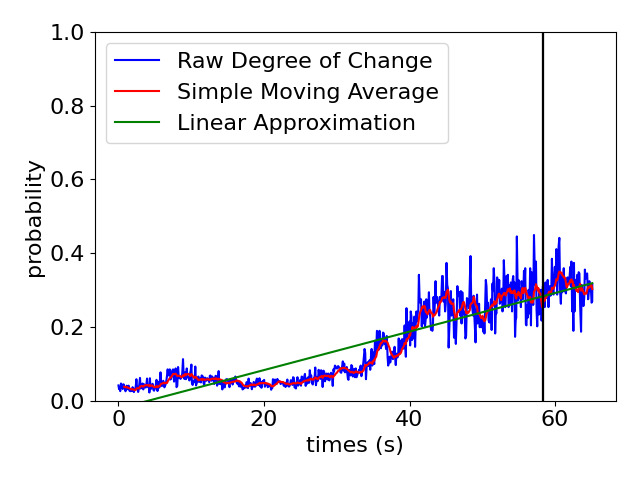}
      \centering (b)-entire
    \end{minipage}
    %% \hspace{0.01\columnwidth}
    \begin{minipage}{0.24\columnwidth}
      \includegraphics[width=\columnwidth]{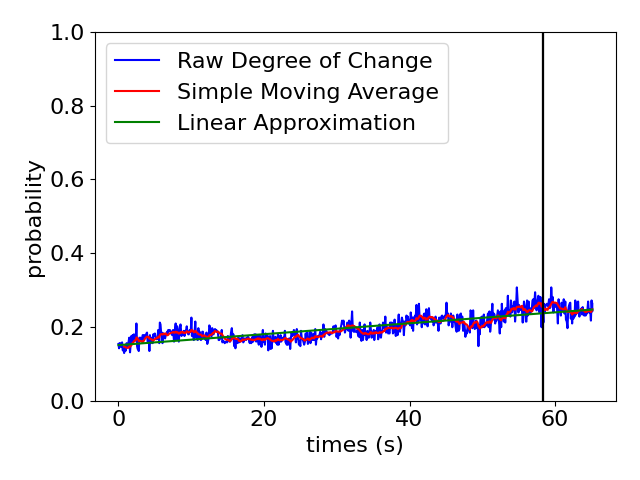}
      \centering (c)-entire
    \end{minipage}
    %% \hspace{0.01\columnwidth}
    \begin{minipage}{0.24\columnwidth}
      \includegraphics[width=\columnwidth]{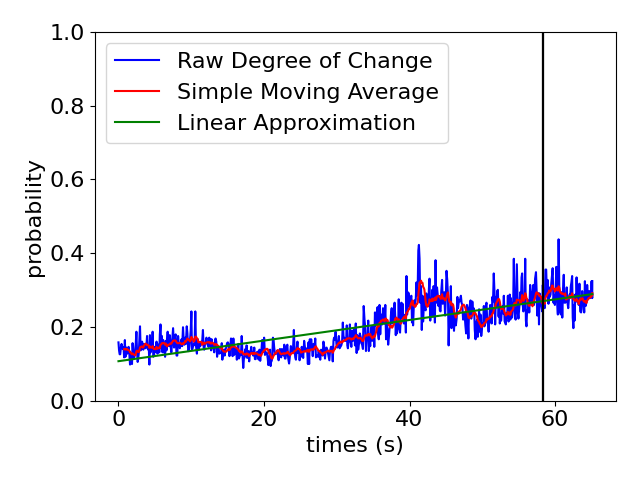}
      \centering (d)-entire
    \end{minipage}
    %% \hspace{0.01\columnwidth}
    \begin{minipage}{0.24\columnwidth}
      \includegraphics[width=\columnwidth]{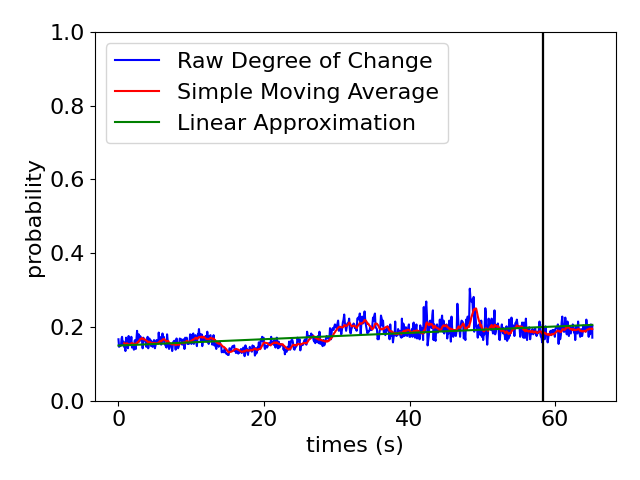}
      \centering (a)-contents
    \end{minipage}
    %% \hspace{0.01\columnwidth}
    \begin{minipage}{0.24\columnwidth}
      \includegraphics[width=\columnwidth]{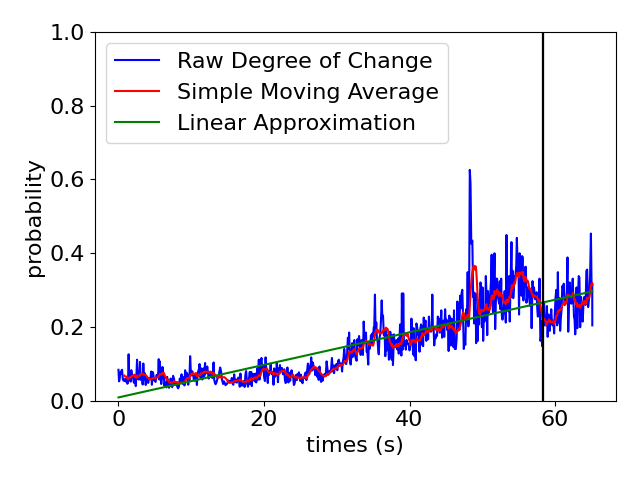}
      \centering (b)-contents
    \end{minipage}
    %% \hspace{0.01\columnwidth}
    \begin{minipage}{0.24\columnwidth}
      \includegraphics[width=\columnwidth]{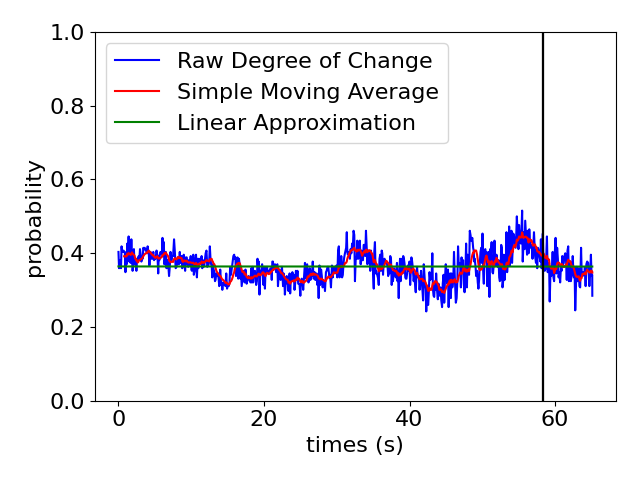}
      \centering (c)-contents
    \end{minipage}
    %% \hspace{0.01\columnwidth}
    \begin{minipage}{0.24\columnwidth}
      \includegraphics[width=\columnwidth]{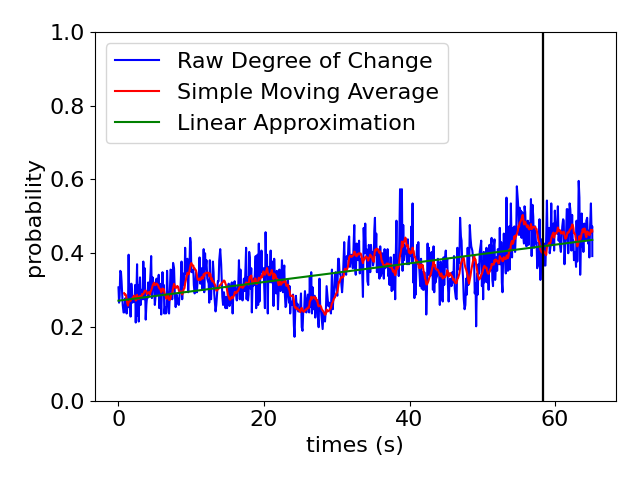}
      \centering (d)-contents
    \end{minipage}
    %% \hspace{0.01\columnwidth}
  \end{center}
  \vspace{-6mm}
  \caption{Plots of inferred degree of state change for each condition against melting data.}
  \vspace{-12mm}
  \label{figure:graph_melting}
\end{figure}

\begin{figure}[h!]
  \begin{center}
    \begin{minipage}{0.48\columnwidth}
      \includegraphics[width=\columnwidth]{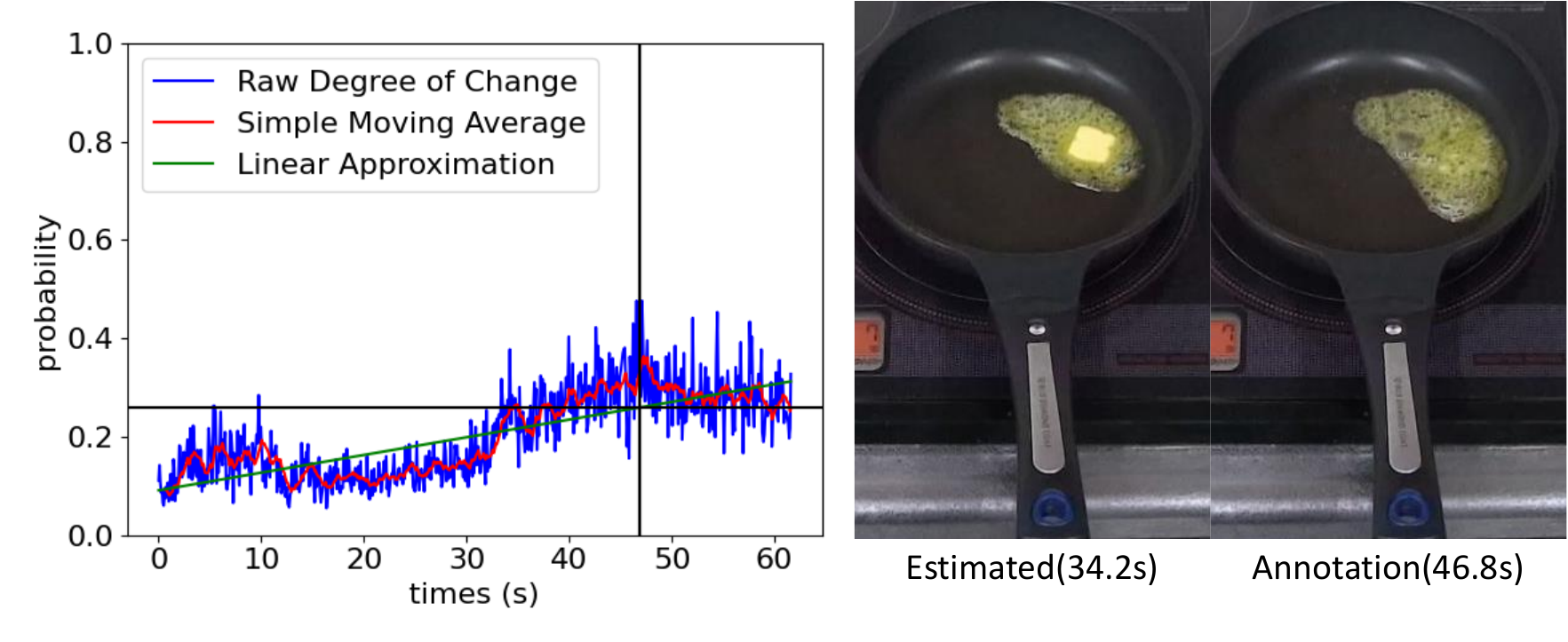}
      \centering same-power-(b)-entire
    \end{minipage}
    %% \hspace{0.01\columnwidth}
    \begin{minipage}{0.48\columnwidth}
      \includegraphics[width=\columnwidth]{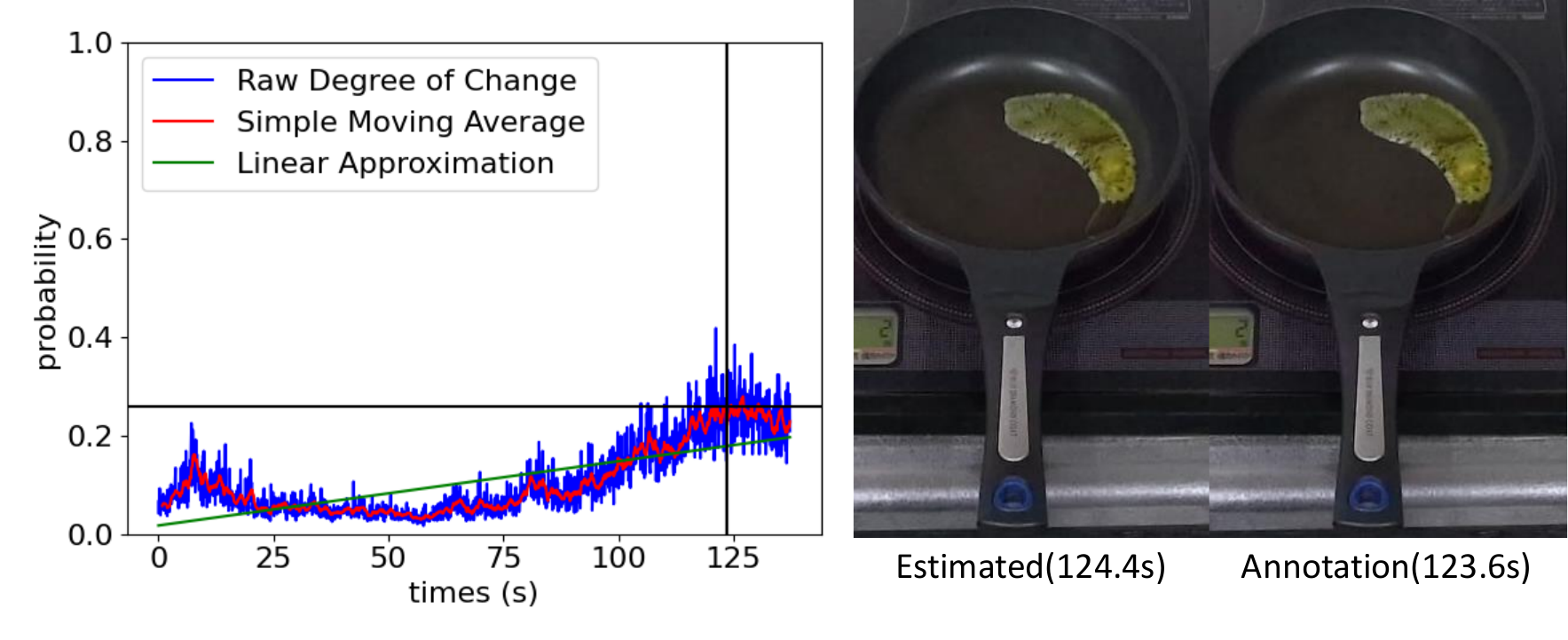}
      \centering different-power-(b)-entire
    \end{minipage}
    %% \hspace{0.01\columnwidth}
    \begin{minipage}{0.48\columnwidth}
      \includegraphics[width=\columnwidth]{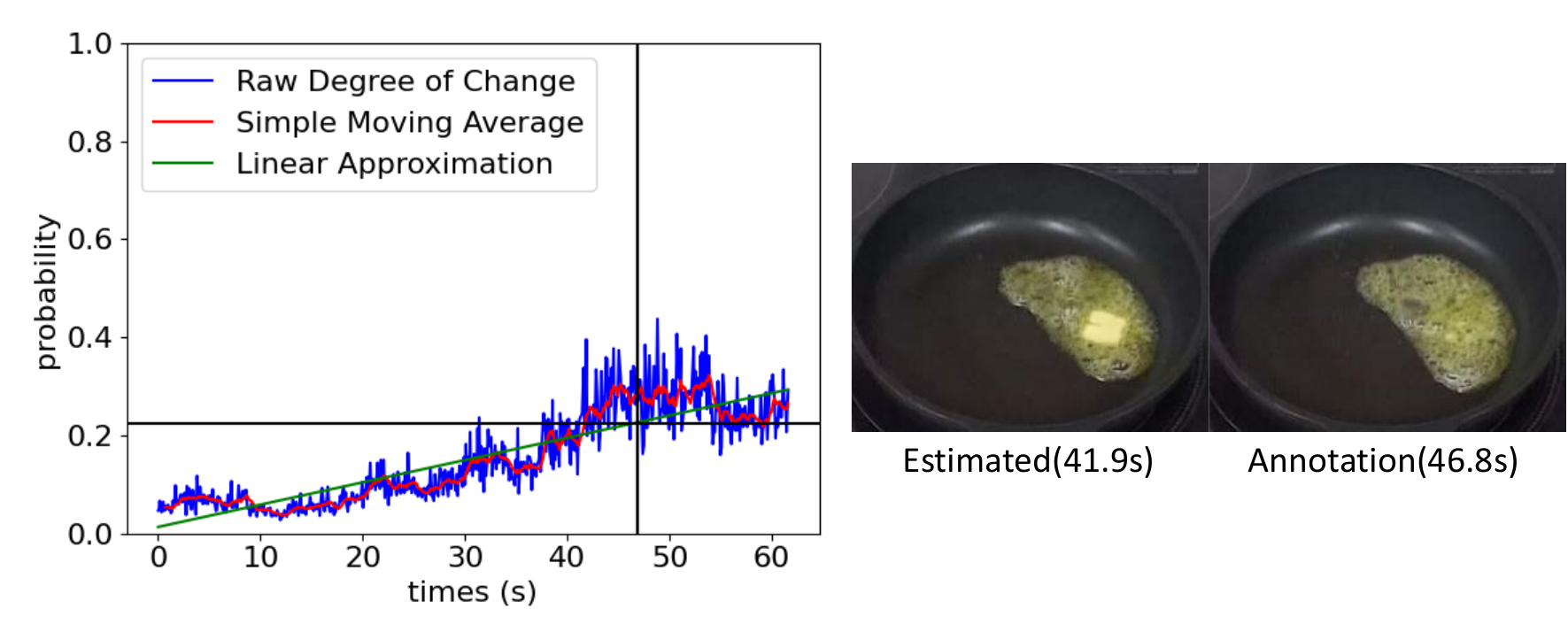}
      \centering same-power-(b)-contents
    \end{minipage}
    %% \hspace{0.01\columnwidth}
    \begin{minipage}{0.48\columnwidth}
      \includegraphics[width=\columnwidth]{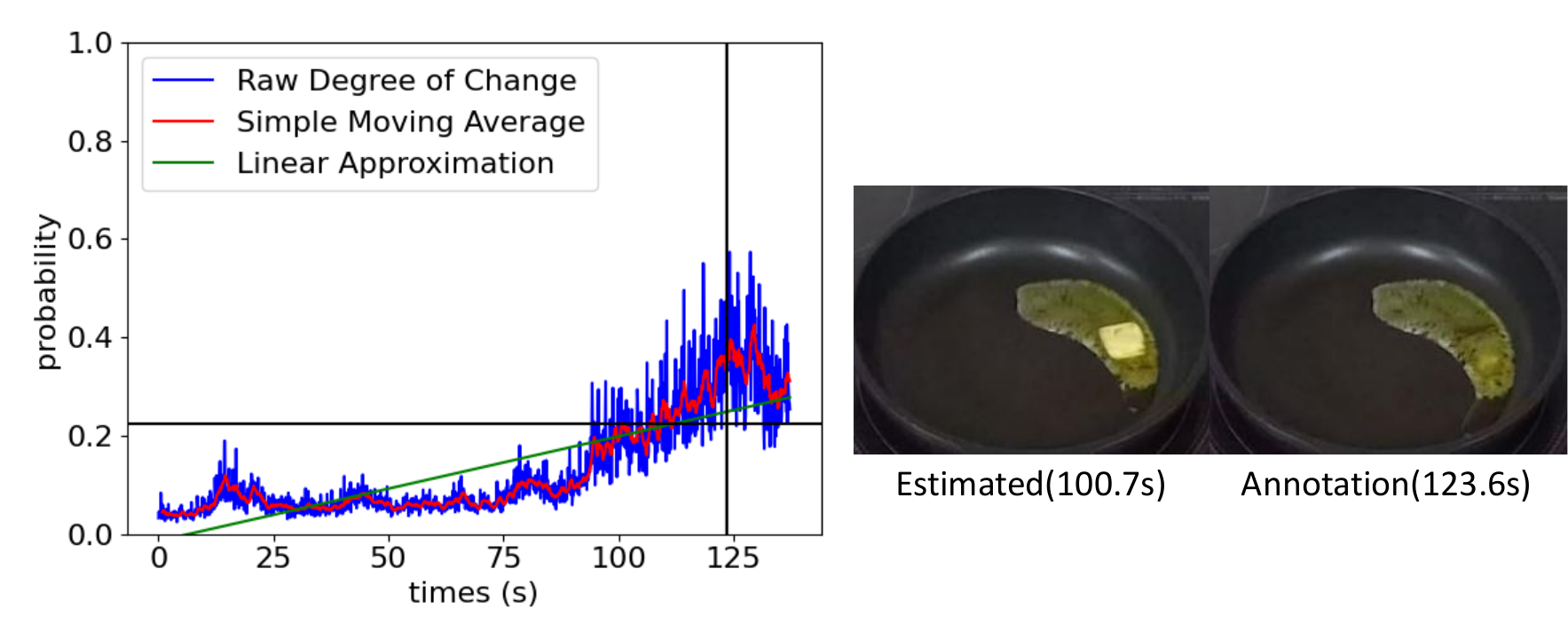}
      \centering different-power-(b)-contents
    \end{minipage}
    %% \hspace{0.01\columnwidth}
  \end{center}
  \vspace{-6mm}
  \caption{Plots of state change recognition results for unknown melting data and comparison of images at the estimated time and at the time annotated by human.}
  \label{figure:result_image_melting}
  \vspace{-10mm}
\end{figure}

For the melting state change, data were collected by cooking butter in a frying pan to melt it (\figref{images_melting}). For melting, four different prompts were designed in the same way, and comparisons were made under each condition. For the butter melting data, prompt (b) was the best for both gazing conditions (\tabref{comparison_melting}, \figref{graph_melting}). The results of the recognizer performance evaluation using the best prompt (b) are shown in \tabref{result_melting} and \figref{result_image_melting}. For the melting data, it was better to look only at the contents of the pan for the same heat level, while it was better to look at the whole pan for the different heat levels.

\begin{table}[h]
  \begin{center}
    \caption{Comparison of prompts and gazing areas for melting data.}
    \vspace{-6mm}
    \label{table:comparison_melting}
    \scalebox{0.8}{
      \begin{tabular}{|c|c|c|c||c|}
        \hline
        & Gaze Area & Positive Prompt & Negative Prompt & LA Slope  \\
        \hline
        (a)-entire & entire pan & Melted butter & Unmelted butter & 0.00183 \\
        (b)-entire & entire pan & Melted liquid butter & Unmelted solid butter & 0.00522 \\
        (c)-entire & entire pan & Butter that has melted & Butter that has not melted & 0.00148 \\
        (d)-entire & entire pan & Butter that has melted and turned to liquid & Butter that has not melted and remains solid & 0.00279 \\
        \hline
        (a)-contents & contents & Melted butter & Unmelted butter & 0.00086 \\
        (b)-contents & contents & Melted liquid butter & Unmelted solid butter & 0.00441 \\
        (c)-contents & contents & Butter that has melted & Butter that has not melted & -0.00000 \\
        (d)-contents & contents & Butter that has melted and turned to liquid & Butter that has not melted and remains solid & 0.00253 \\
        \hline
    \end{tabular}}
    \vspace{-12mm}
  \end{center}
\end{table}

\begin{table}[h!]
  \begin{center}
    \caption{State change recognition results for unknown data of melting.}
    \vspace{-3mm}
    \label{table:result_melting}
    \scalebox{0.9}{
      \begin{tabular}{|c||c|c|}
        \hline
        & Same Power Diff (s) &  Different Power Diff (s)  \\
        \hline
        (b)-entire & 12.6 & 0.9 \\
        \hline
        (b)-contents & 5.0 & 22.9 \\
        \hline
    \end{tabular}}
    \vspace{-12mm}
  \end{center}
\end{table}

\subsubsection{Thermal Denaturation of Proteins.}
%% タンパク質の熱変性については，フライパンで卵を加熱し凝固させる調理で，目玉焼きを作るような調理を行ってデータを収集した(figref{})．同様に4種類のプロンプトを比較を行った結果，どちらの注視条件においても(b)のプロンプトが最良であるという結果になった．そのプロンプトを用いた状態変化認識器性能評価の結果を\tabref{result}に示す．異なる火力における鍋全体を注視した条件以外については，認識器としてうまく機能していないという結果になった．

\begin{figure}[htb]
  \begin{center}
    \begin{minipage}{0.085\columnwidth}
      \includegraphics[height=1.6cm]{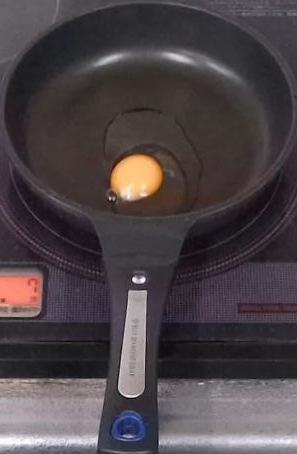}
      \centering 0s
    \end{minipage}
    %% \hspace{0.01\columnwidth}
    \begin{minipage}{0.085\columnwidth}
      \includegraphics[height=1.6cm]{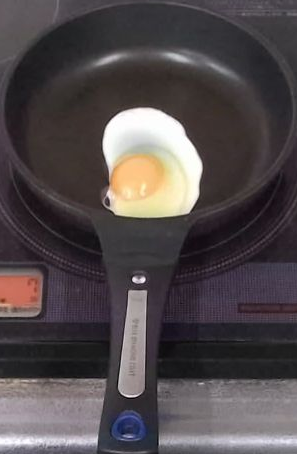}
      \centering 30s
    \end{minipage}
    %% \hspace{0.01\columnwidth}
    \begin{minipage}{0.085\columnwidth}
      \includegraphics[width=\columnwidth]{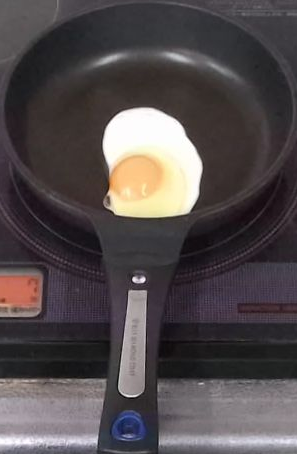}
      \centering 60s
    \end{minipage}
    %% \hspace{0.01\columnwidth}
    \begin{minipage}{0.085\columnwidth}
      \includegraphics[height=1.6cm]{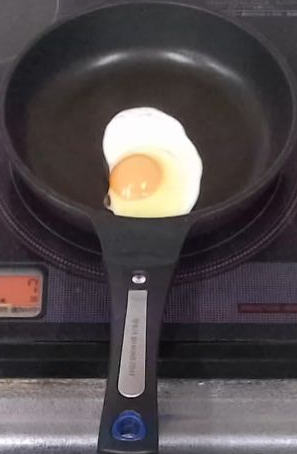}
      \centering 90s
    \end{minipage}
    %% \hspace{0.01\columnwidth}
    \begin{minipage}{0.085\columnwidth}
      \includegraphics[height=1.6cm]{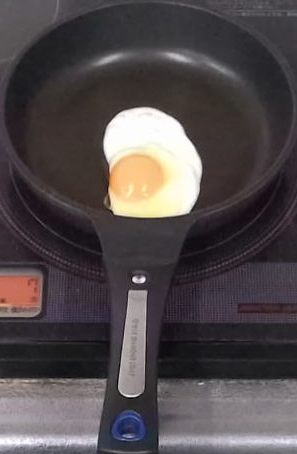}
      \centering 120s
    \end{minipage}
    %% \hspace{0.01\columnwidth}
    \begin{minipage}{0.085\columnwidth}
      \includegraphics[height=1.6cm]{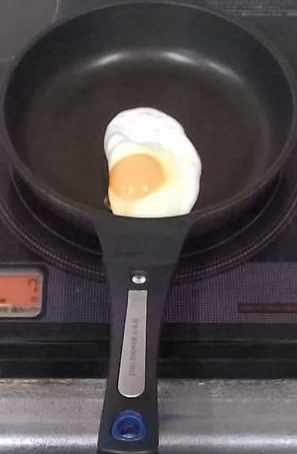}
      \centering 150s
    \end{minipage}
    %% \hspace{0.01\columnwidth}
    \begin{minipage}{0.085\columnwidth}
      \includegraphics[height=1.6cm]{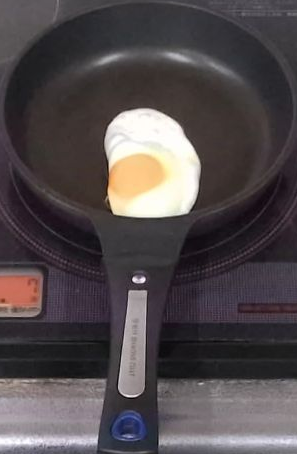}
      \centering 180s
    \end{minipage}
    \hspace{0.01\columnwidth}
    \begin{minipage}{0.085\columnwidth}
      \includegraphics[height=1.6cm]{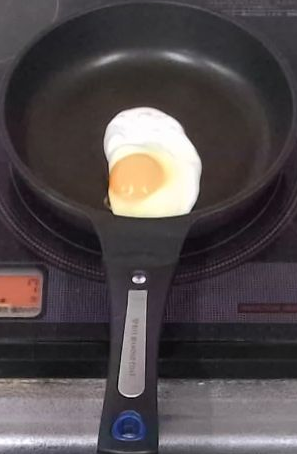}
      \centering (A)
    \end{minipage}
    %% \hspace{0.01\columnwidth}
    \begin{minipage}{0.16\columnwidth}
      \includegraphics[height=1.6cm]{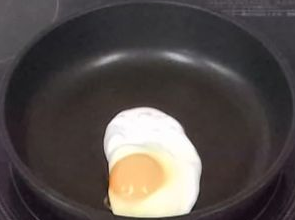}
      \centering (B)
    \end{minipage}
    %% \hspace{0.01\columnwidth}
  \end{center}
  \vspace{-6mm}
  \caption{State change of thermal denaturation of proteins. (A) Image of the entire pot at the time when the person felt the state change occurred (139.0s). (B) Image of only the contents of the pot at the same time.}
  \label{figure:images_denaturation}
  \vspace{-12mm}
\end{figure}

\begin{figure}[h!]
  \begin{center}
    \begin{minipage}{0.24\columnwidth}
      \includegraphics[width=\columnwidth]{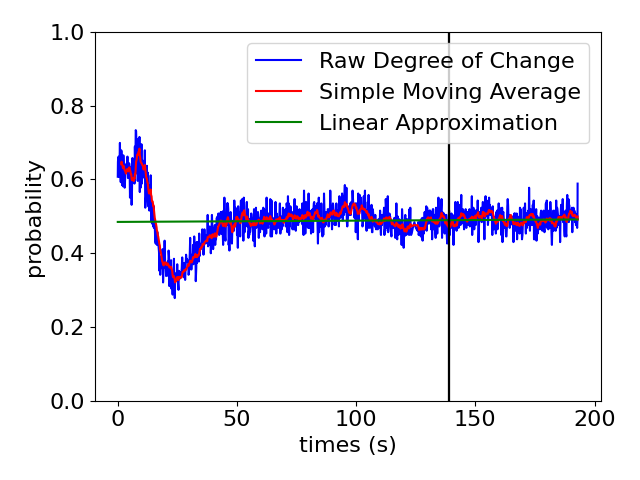}
      \centering (a)-entire
    \end{minipage}
    %% \hspace{0.01\columnwidth}
    \begin{minipage}{0.24\columnwidth}
      \includegraphics[width=\columnwidth]{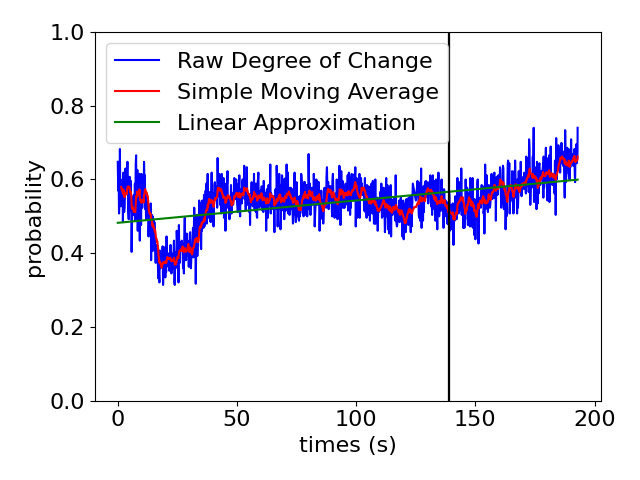}
      \centering (b)-entire
    \end{minipage}
    %% \hspace{0.01\columnwidth}
    \begin{minipage}{0.24\columnwidth}
      \includegraphics[width=\columnwidth]{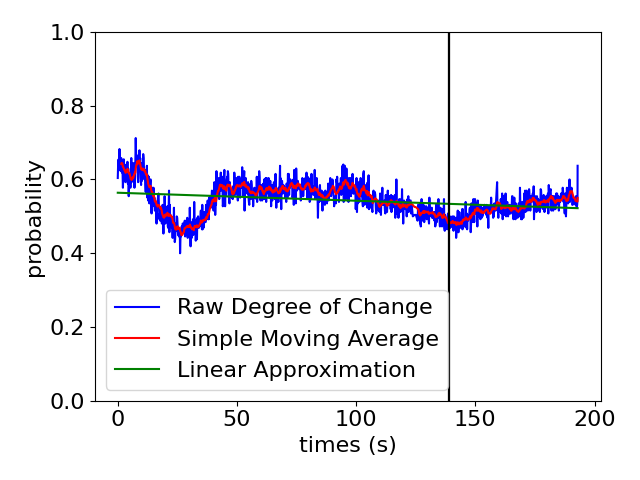}
      \centering (c)-entire
    \end{minipage}
    %% \hspace{0.01\columnwidth}
    \begin{minipage}{0.24\columnwidth}
      \includegraphics[width=\columnwidth]{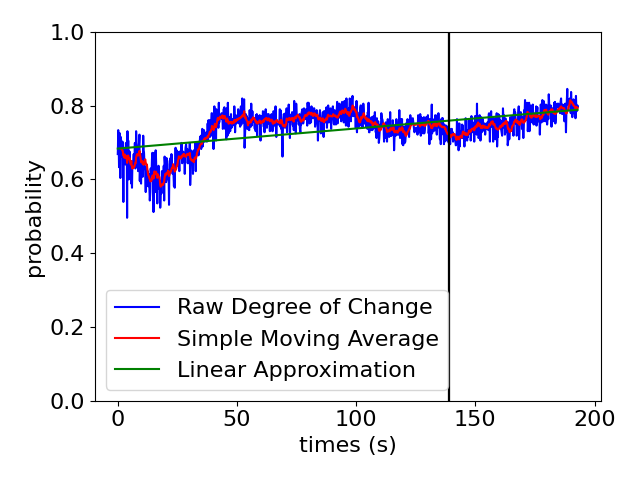}
      \centering (d)-entire
    \end{minipage}
    %% \hspace{0.01\columnwidth}
    \begin{minipage}{0.24\columnwidth}
      \includegraphics[width=\columnwidth]{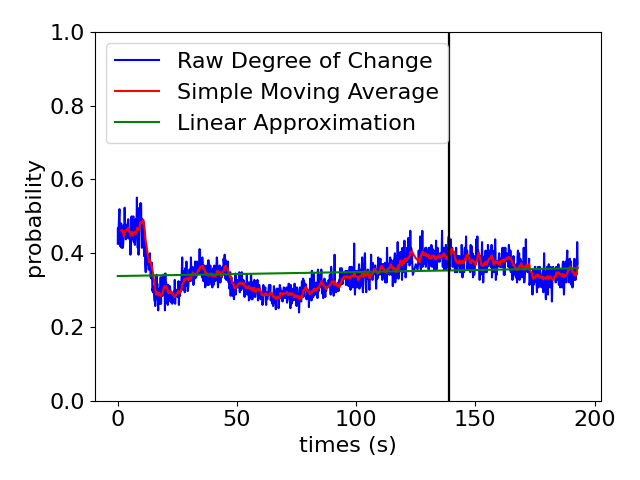}
      \centering (a)-contents
    \end{minipage}
    %% \hspace{0.01\columnwidth}
    \begin{minipage}{0.24\columnwidth}
      \includegraphics[width=\columnwidth]{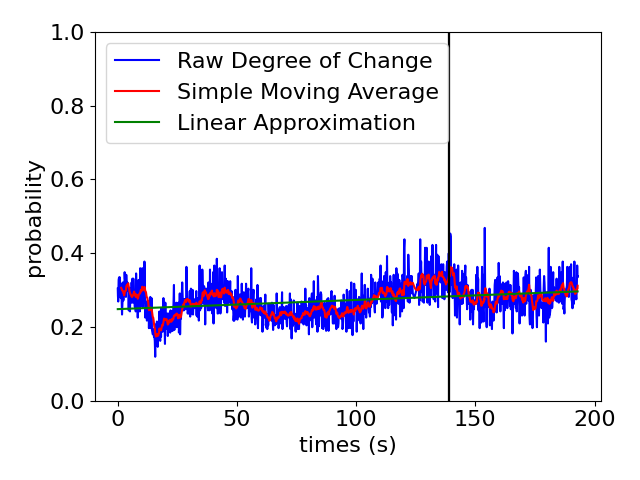}
      \centering (b)-contents
    \end{minipage}
    %% \hspace{0.01\columnwidth}
    \begin{minipage}{0.24\columnwidth}
      \includegraphics[width=\columnwidth]{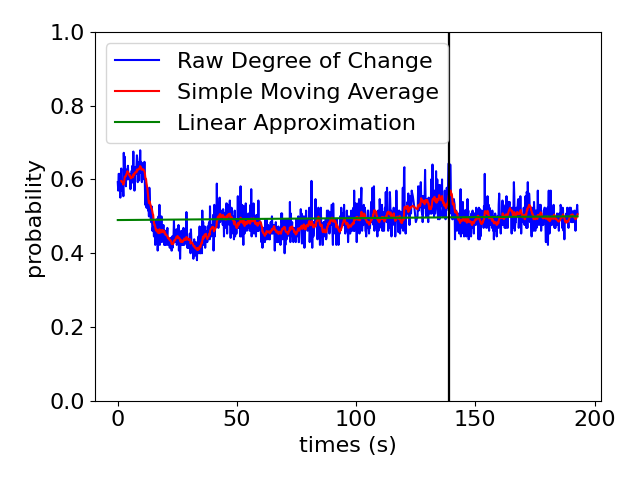}
      \centering (c)-contents
    \end{minipage}
    %% \hspace{0.01\columnwidth}
    \begin{minipage}{0.24\columnwidth}
      \includegraphics[width=\columnwidth]{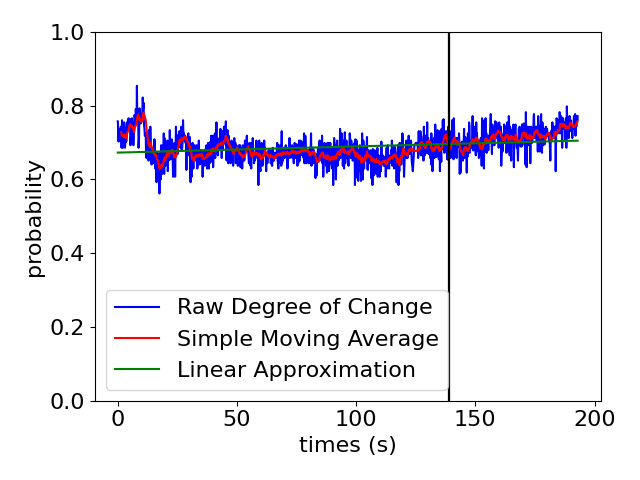}
      \centering (d)-contents
    \end{minipage}
    %% \hspace{0.01\columnwidth}
  \end{center}
  \vspace{-6mm}
  \caption{Plots of inferred degree of state change for each condition against thermal denaturation of proteins data}
  \label{figure:graph_denaturation}
  \vspace{-6mm}
\end{figure}

\begin{figure}[h!]
  \begin{center}
    \begin{minipage}{0.48\columnwidth}
      \includegraphics[width=\columnwidth]{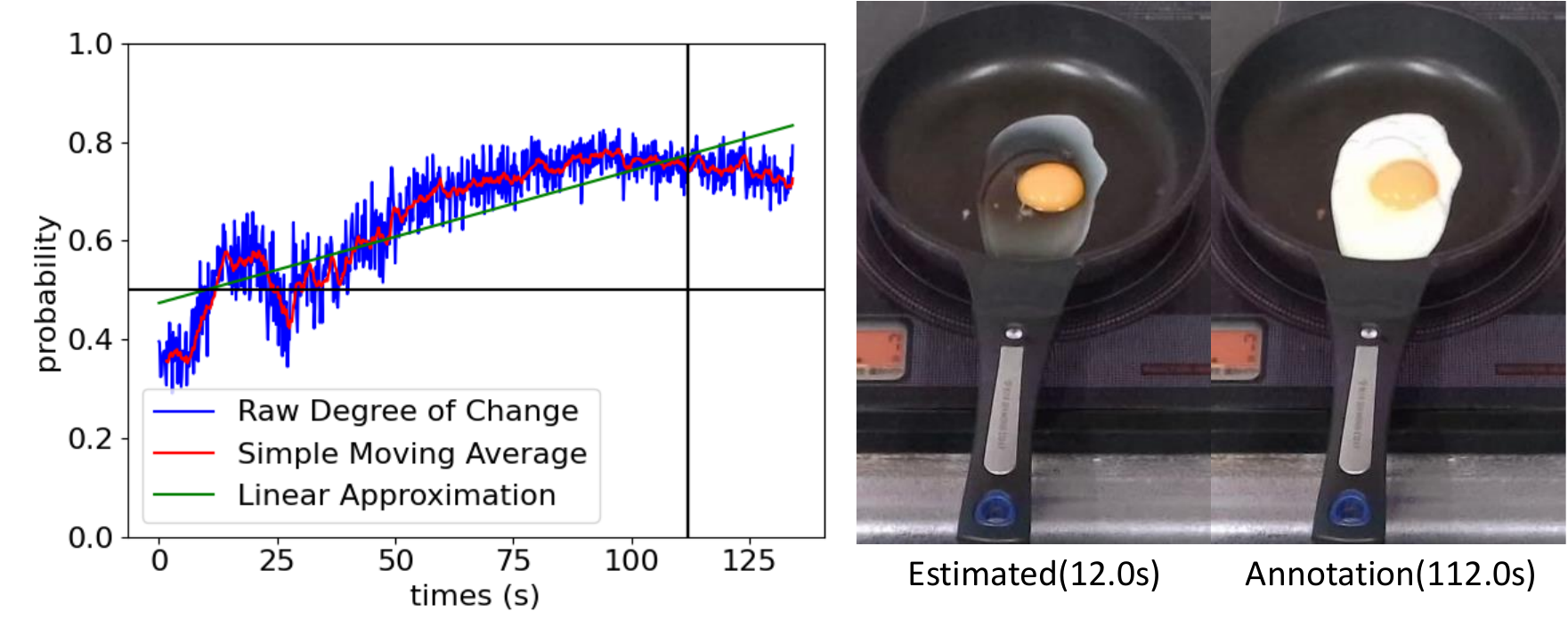}
      \centering same-power-(b)-entire
    \end{minipage}
    %% \hspace{0.01\columnwidth}
    \begin{minipage}{0.48\columnwidth}
      \includegraphics[width=\columnwidth]{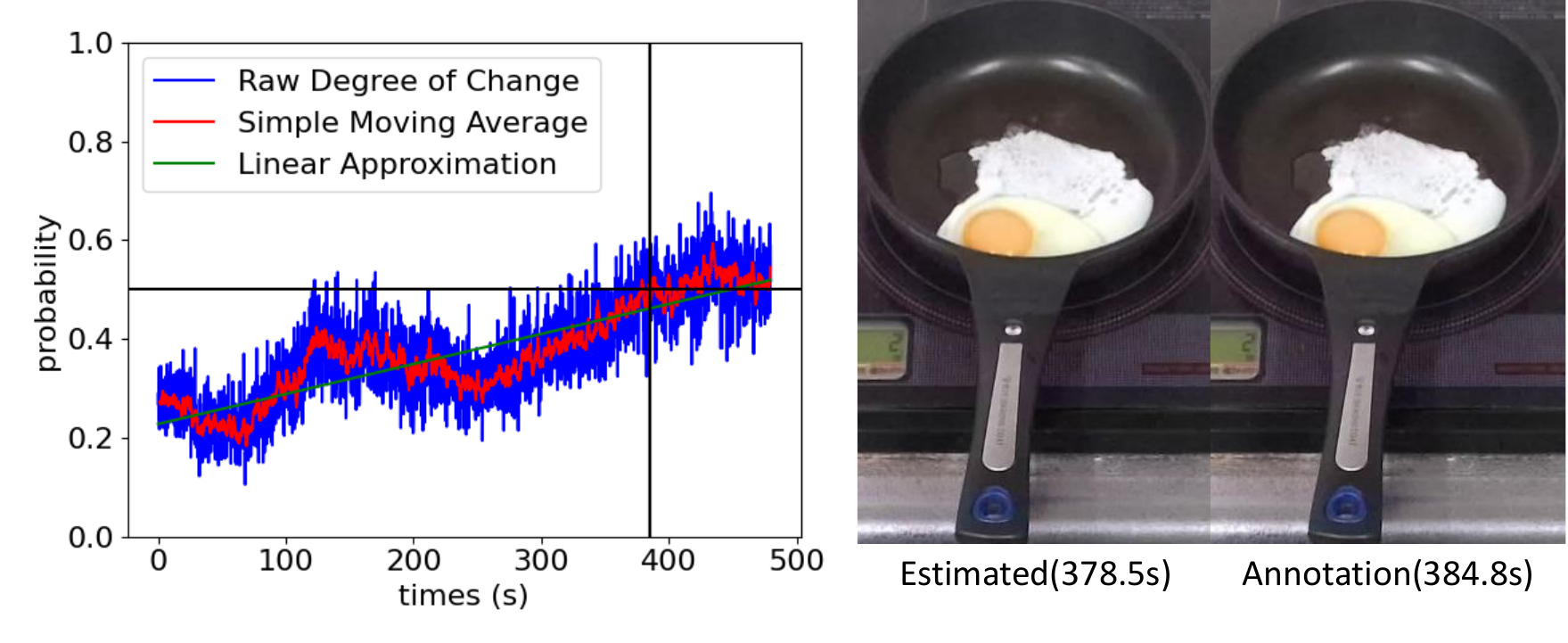}
      \centering different-power-(b)-entire
    \end{minipage}
    %% \hspace{0.01\columnwidth}
    \begin{minipage}{0.48\columnwidth}
      \includegraphics[width=\columnwidth]{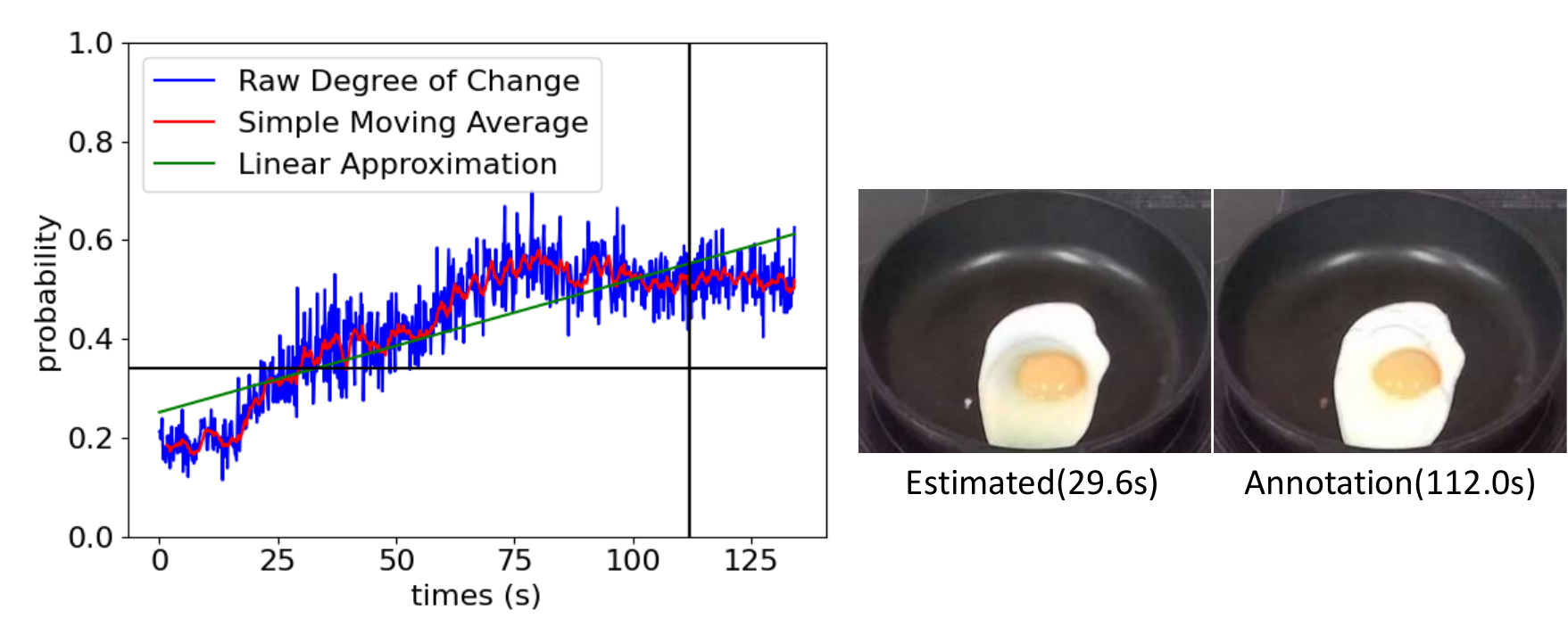}
      \centering same-power-(b)-contents
    \end{minipage}
    %% \hspace{0.01\columnwidth}
    \begin{minipage}{0.48\columnwidth}
      \includegraphics[width=\columnwidth]{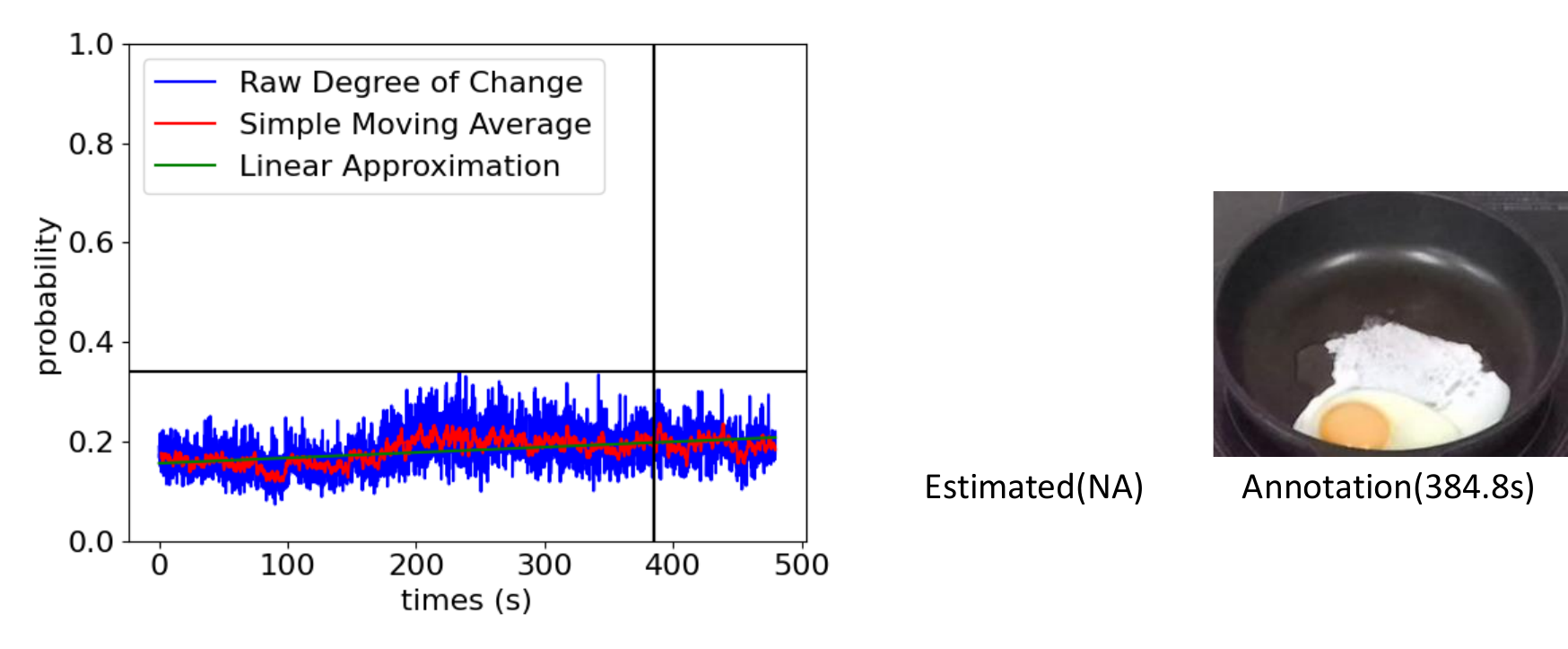}
      \centering different-power-(b)-contents
    \end{minipage}
    %% \hspace{0.01\columnwidth}
  \end{center}
  \vspace{-6mm}
  \caption{Plots of state change recognition results for unknown thermal denaturation of proteins data and comparison of images at the estimated time and at the time annotated by human.}
  \label{figure:result_image_denaturation}
  \vspace{-6mm}
\end{figure}

For the thermal denaturation of proteins, data were collected by cooking eggs in a frying pan to coagulate them, similar to making sunny-side up (\figref{images_denaturation}). Similarly, four types of prompts were compared, with the result that prompt (b) was the best under both gazing conditions (\tabref{comparison_denaturation}, \figref{graph_denaturation}). The results of the state change recognizer performance evaluation using these prompts are shown in \tabref{result_denaturation} and \figref{result_image_denaturation}. The results showed that the system did not work well as a recognizer for all conditions except for the condition in which the entire pot was gazed at at different thermal powers.

\begin{table}[h!]
  \begin{center}
    \caption{Comparison of prompts and gazing areas for thermal denaturation of proteins data.}
    \vspace{-6mm}
    \label{table:comparison_denaturation}
    \scalebox{0.8}{
      \begin{tabular}{|c|c|c|c||c|}
        \hline
        & Gaze Area & Positive Prompt & Negative Prompt & LA Slope  \\
        \hline
        (a)-entire & entire pan & Cooked egg & Uncooked egg & 0.00003 \\
        (b)-entire & entire pan & Cooked and whitened egg & Uncooked and raw egg & 0.00061 \\
        (c)-entire & entire pan & Egg that has been cooked & Egg that has not been cooked & -0.00022 \\
        (d)-entire & entire pan & Egg that has been cooked and turned white & Egg that has not been cooked and is raw & 0.00055 \\
        \hline
        (a)-contents & contents & Cooked egg & Uncooked egg & 0.00011 \\
        (b)-contents & contents & Cooked and whitened egg & Uncooked and raw egg & 0.00025 \\
        (c)-contents & contents & Egg that has been cooked & Egg that has not been cooked & 0.00005 \\
        (d)-contents & contents & Egg that has been cooked and turned white & Egg that has not been cooked and is raw & 0.00017 \\
        \hline
    \end{tabular}}
    \vspace{-16mm}
  \end{center}
\end{table}

\begin{table}[h]
  \begin{center}
    \caption{State change recognition results for unknown data of thermal denaturation of proteins.}
    \vspace{-3mm}
    \label{table:result_denaturation}
    \scalebox{0.9}{
      \begin{tabular}{|c||c|c|}
        \hline
        & Same Power Diff (s) &  Different Power Diff (s)  \\
        \hline
        (b)-entire & 100.0 & 6.3 \\
        \hline
        (b)-contents & 82.3 & NA \\
        \hline
    \end{tabular}}
    \vspace{-6mm}
  \end{center}
\end{table}

The plots of the graphs show that even when the gazing conditions and prompts are the same, there are differences in the way the plots are drawn for each data, indicating that the recognition is not successful. The reasons for this may include the possibility that the four types of prompts used in this study do not capture the state change well, or that the thermal denaturation of proteins is a state change with a larger visual difference from one data to another than other state changes.

\subsubsection{Maillard Reaction.}

%% メイラード反応の実験として，フライパンで玉ねぎを飴色に炒める調理のデータ収集を行った(\figref{})．玉ねぎを炒める際には，ヘラで玉ねぎをかき混ぜないとすぐに焦げてしまうため，ダイレクトティーチした動作によりロボットが玉ねぎをかき混ぜながら炒めた．ロボットの腕やヘラが映り込むと状態推定が非常に不安定になってしまうため，ロボットは定期的にかき混ぜる動作を停止して玉ねぎの画像を保存するようにした．そのため，これまでの3つの状態変化とは異なり離散的なデータが収集される．離散的な少数のデータであるため，閾値処理においても単純移動平均は行わずに生の状態変化度の値を用いて認識した．

%% これまでの3つと同様の方式で4種類のプロンプトを用意し，比較を行った．どちらの注視条件においても(b)のプロンプトが最適だと選択され，未知データに対する状態変化認識を行った．どちらの火力条件においても，鍋全体を注視する条件においてアノテーションと推論時刻が一致し，鍋の中身のみを注視するよりも良い結果となった．ピッタリと一致したのは離散的なデータであるため他の3つとくらべて判定される時刻の候補が少ないからである．

\begin{figure}[h!]
  \begin{center}
    \vspace{-6mm}
    \begin{minipage}{0.088\columnwidth}
      \includegraphics[height=1.6cm]{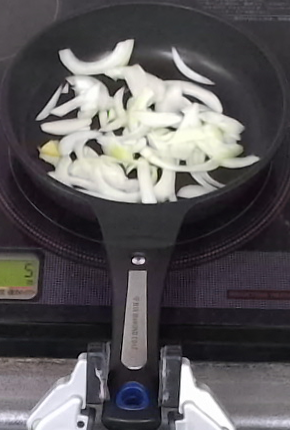}
      \centering 0s
    \end{minipage}
    %% \hspace{0.01\columnwidth}
    \begin{minipage}{0.088\columnwidth}
      \includegraphics[height=1.6cm]{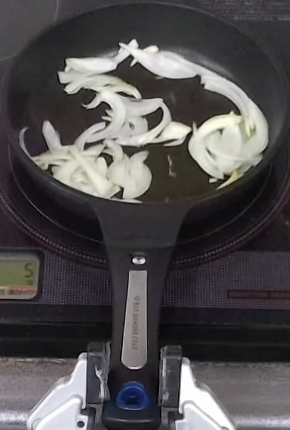}
      \centering 170s
    \end{minipage}
    %% \hspace{0.01\columnwidth}
    \begin{minipage}{0.088\columnwidth}
      \includegraphics[height=1.6cm]{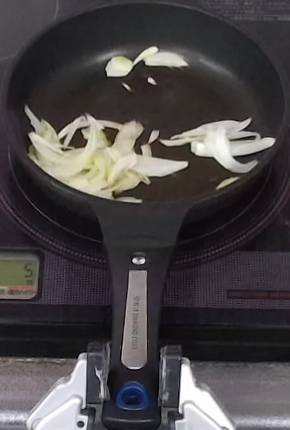}
      \centering 339s
    \end{minipage}
    %% \hspace{0.01\columnwidth}
    \begin{minipage}{0.088\columnwidth}
      \includegraphics[height=1.6cm]{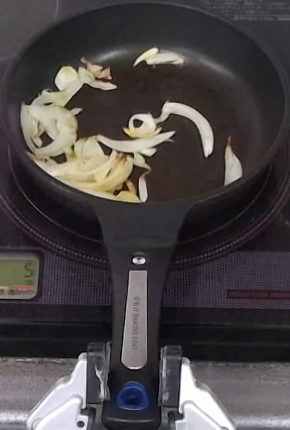}
      \centering 510s
    \end{minipage}
    %% \hspace{0.01\columnwidth}
    \begin{minipage}{0.088\columnwidth}
      \includegraphics[height=1.6cm]{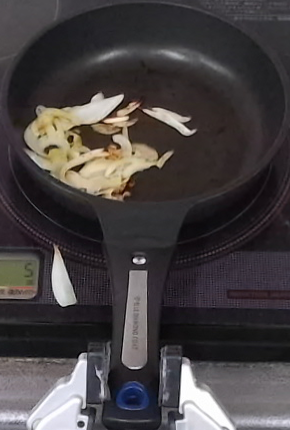}
      \centering 681s
    \end{minipage}
    %% \hspace{0.01\columnwidth}
    \begin{minipage}{0.088\columnwidth}
      \includegraphics[height=1.6cm]{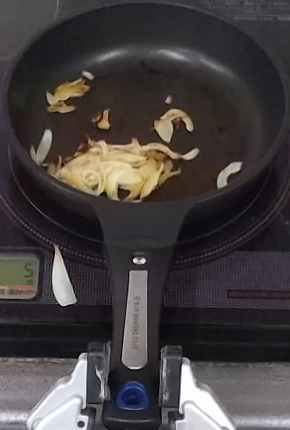}
      \centering 852s
    \end{minipage}
    %% \hspace{0.01\columnwidth}
    \begin{minipage}{0.088\columnwidth}
      \includegraphics[height=1.6cm]{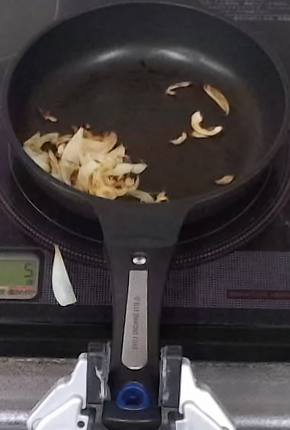}
      \centering 1022s
    \end{minipage}
    \hspace{0.01\columnwidth}
    \begin{minipage}{0.088\columnwidth}
      \includegraphics[height=1.6cm]{figs/20220212_cook_onion_pan/20220212_onion_02_15_ano_human_pan_1676205070661295532}
      \centering (A)
    \end{minipage}
    %% \hspace{0.01\columnwidth}
    \begin{minipage}{0.16\columnwidth}
      \includegraphics[height=1.6cm]{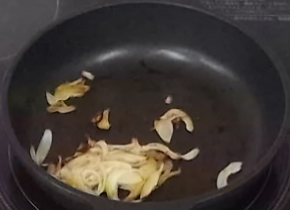}
      \centering (B)
    \end{minipage}
    %% \hspace{0.01\columnwidth}
  \end{center}
  \vspace{-6mm}
  \caption{State change of Maillard reaction. (A) Image of the entire pot at the time when the person felt the state change occurred (851.6s). (B) Image of only the contents of the pot at the same time.}
  \vspace{-6mm}
  \label{figure:images_maillard}
\end{figure}

As an experiment of Maillard reaction, we collected data of cooking onions in a frying pan to fry them to a candy color (\figref{images_maillard}). When frying onions, the robot stir-fried the onions with a direct-teach motion because the onions would burn quickly if they were not stirred with a spatula. The robot periodically stopped its stirring motion to save the image of the onion, because the reflection of the robot's arm or the spatula would make the state estimation very unstable. Therefore, unlike the previous three state changes, discrete data is collected. Because of the small number of discrete data, no simple moving average was used in the thresholding process, but the raw state change values were used for recognition.

\begin{figure}[h!]
  \begin{center}
    \begin{minipage}{0.24\columnwidth}
      \includegraphics[width=\columnwidth]{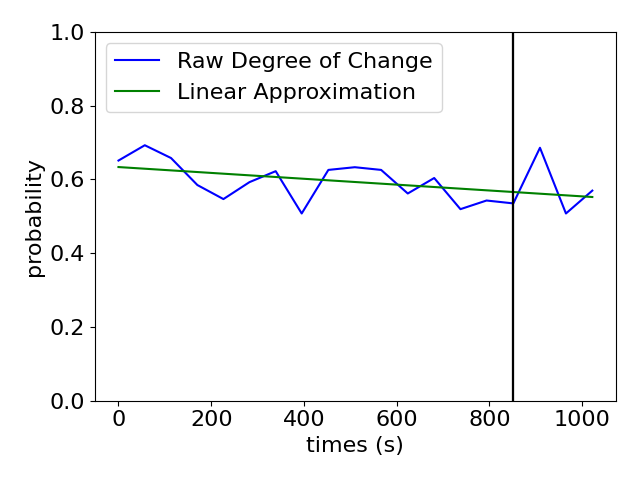}
      \centering (a)-entire
    \end{minipage}
    %% \hspace{0.01\columnwidth}
    \begin{minipage}{0.24\columnwidth}
      \includegraphics[width=\columnwidth]{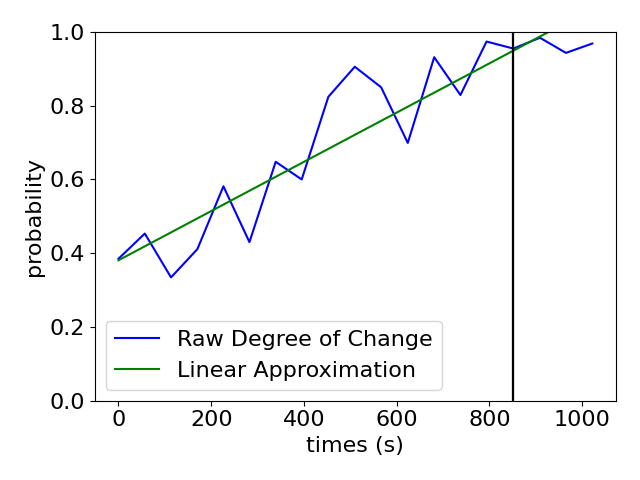}
      \centering (b)-entire
    \end{minipage}
    %% \hspace{0.01\columnwidth}
    \begin{minipage}{0.24\columnwidth}
      \includegraphics[width=\columnwidth]{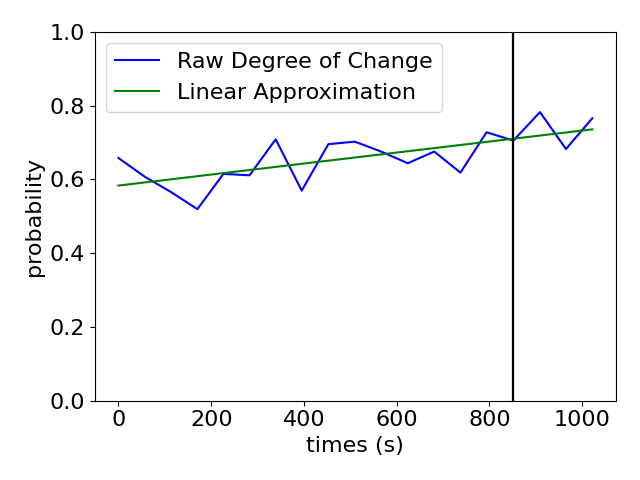}
      \centering (c)-entire
    \end{minipage}
    %% \hspace{0.01\columnwidth}
    \begin{minipage}{0.24\columnwidth}
      \includegraphics[width=\columnwidth]{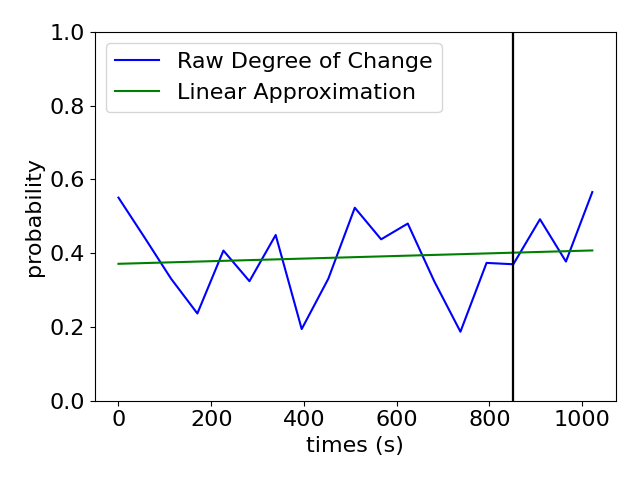}
      \centering (d)-entire
    \end{minipage}
    %% \hspace{0.01\columnwidth}
    \begin{minipage}{0.24\columnwidth}
      \includegraphics[width=\columnwidth]{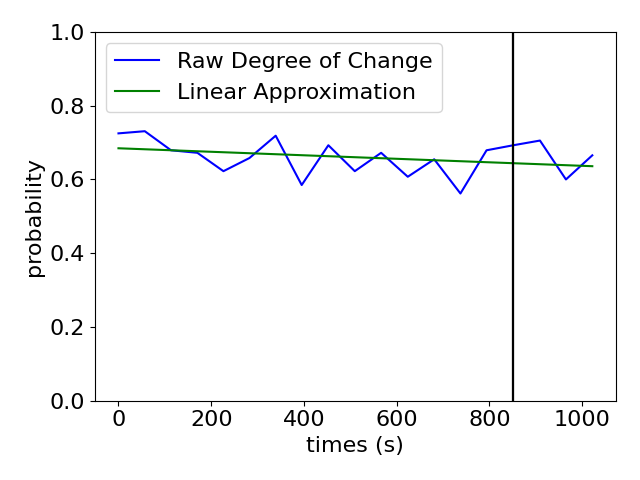}
      \centering (a)-contents
    \end{minipage}
    %% \hspace{0.01\columnwidth}
    \begin{minipage}{0.24\columnwidth}
      \includegraphics[width=\columnwidth]{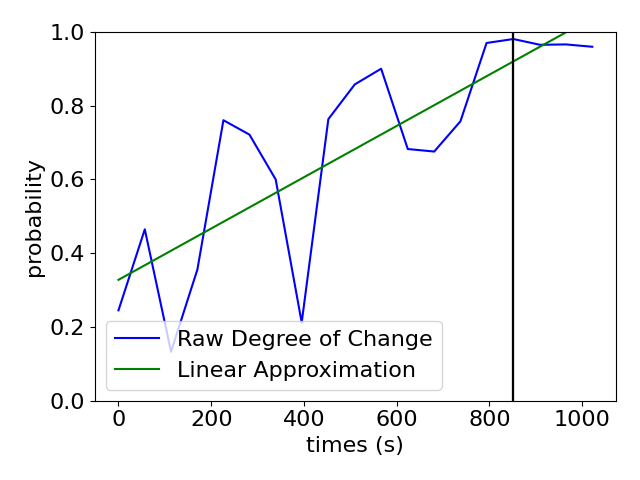}
      \centering (b)-contents
    \end{minipage}
    %% \hspace{0.01\columnwidth}
    \begin{minipage}{0.24\columnwidth}
      \includegraphics[width=\columnwidth]{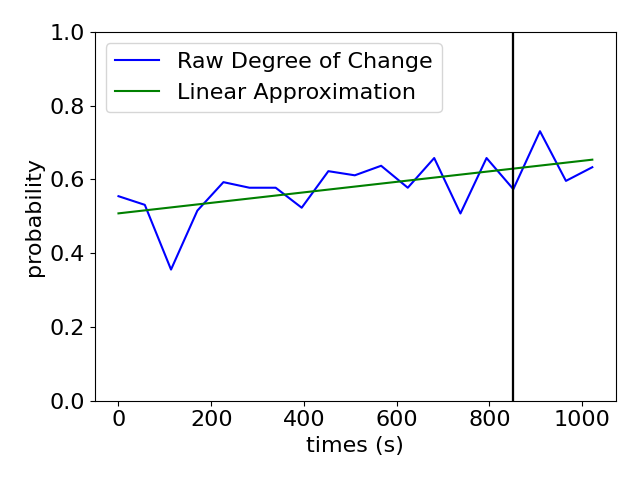}
      \centering (c)-contents
    \end{minipage}
    %% \hspace{0.01\columnwidth}
    \begin{minipage}{0.24\columnwidth}
      \includegraphics[width=\columnwidth]{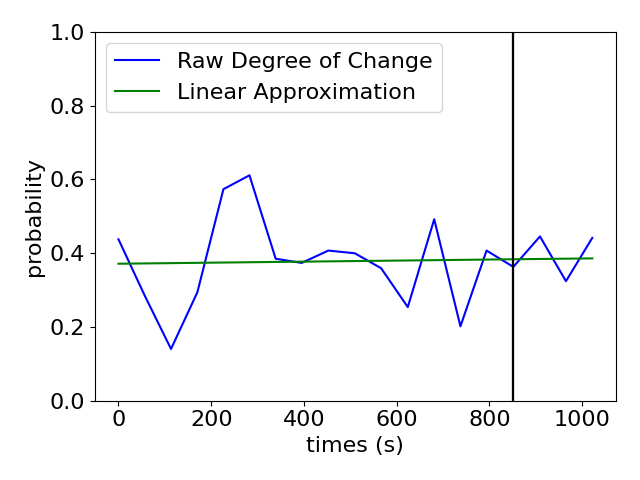}
      \centering (d)-contents
    \end{minipage}
    %% \hspace{0.01\columnwidth}
  \end{center}
  \vspace{-6mm}
  \caption{Plots of inferred degree of state change for each condition against thermal denaturation of proteins data}
  \label{figure:graph_maillard}
  \vspace{-12mm}
\end{figure}

Four different prompts were prepared and compared in the same manner as the previous three (\tabref{comparison_maillard}, \figref{graph_maillard}). Prompt (b) was selected as the best for both gazing conditions, and it was used to recognize the state change for the unknown data (\tabref{result_denaturation}, \figref{result_image_denaturation}). In both thermal conditions, the annotations and estimated times matched in the condition in which the entire pan was gazed at, which was better than when only the contents were gazed at. The reason for the perfect matching is that the data are discrete, so there are fewer candidate times to be judged than in the other three cases.

\begin{table}[h]
  \begin{center}
    \vspace{-3mm}
    \caption{Comparison of prompts and gazing areas for Maillard reaction data.}
    \vspace{-3mm}
    \label{table:comparison_maillard}
    \scalebox{0.7}{
      \begin{tabular}{|c|c|c|c||c|}
        \hline
        & Gaze Area & Positive Prompt & Negative Prompt & LA Slope  \\
        \hline
        (a)-entire & entire pan & Sauteed onions & Unsauteed onions & 0.00008 \\
        (b)-entire & entire pan & Sauteed and candied onions & Still raw and white onions & 0.00067 \\
        (c)-entire & entire pan & Onions that have been sauteed & Onions that have not been sauteed & 0.00015 \\
        (d)-entire & entire pan & Onions that have been sauteed and candied & Onions that have not been sauteed and are still raw & 0.00004 \\
        \hline
        (a)-contents & contents & Sauteed onions & Unsauteed onions & -0.00005 \\
        (b)-contents & contents & Sauteed and candied onions & Still raw and white onions & 0.00070 \\
        (c)-contents & contents & Onions that have been sauteed & Onions that have not been sauteed & 0.00001 \\
        (d)-contents & contents & Onions that have been sauteed and candied & Onions that have not been sauteed and are still raw & 0.00014 \\
        \hline
    \end{tabular}}
    \vspace{-16mm}
  \end{center}
\end{table}

\begin{table}[h!]
  \begin{center}
    \caption{State change recognition results for unknown data of Maillard reaction}
  \vspace{-3mm}
    \label{table:result_maillard}
    \scalebox{0.9}{
      \begin{tabular}{|c||c|c|}
        \hline
        & Same Power Diff (s) &  Different Power Diff (s)  \\
        \hline
        (b)-entire & 0.0 & 0.0 \\
        \hline
        (b)-contents & 113.4 & 107.6 \\
        \hline
    \end{tabular}}
    \vspace{-14mm}
  \end{center}
\end{table}

\begin{figure}[h!]
  \begin{center}
    \begin{minipage}{0.48\columnwidth}
      \includegraphics[width=\columnwidth]{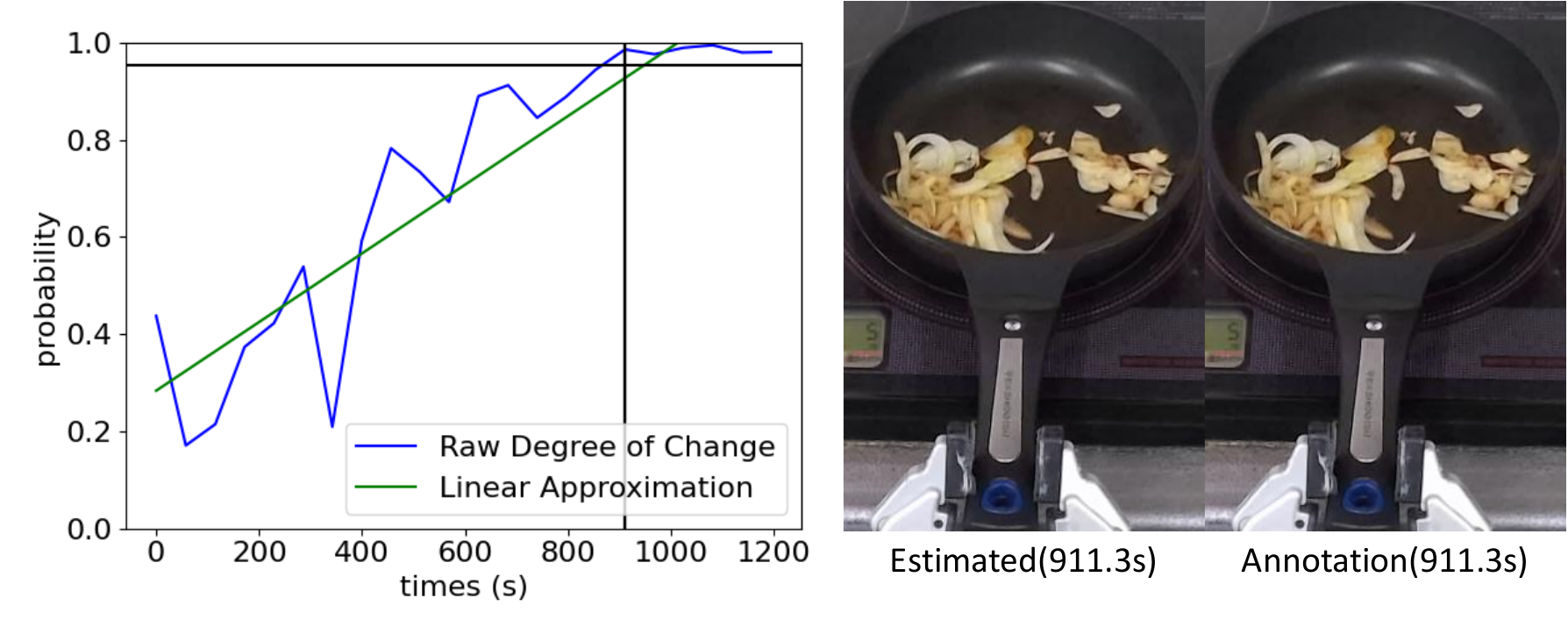}
      \centering same-power-(b)-entire
    \end{minipage}
    %% \hspace{0.01\columnwidth}
    \begin{minipage}{0.48\columnwidth}
      \includegraphics[width=\columnwidth]{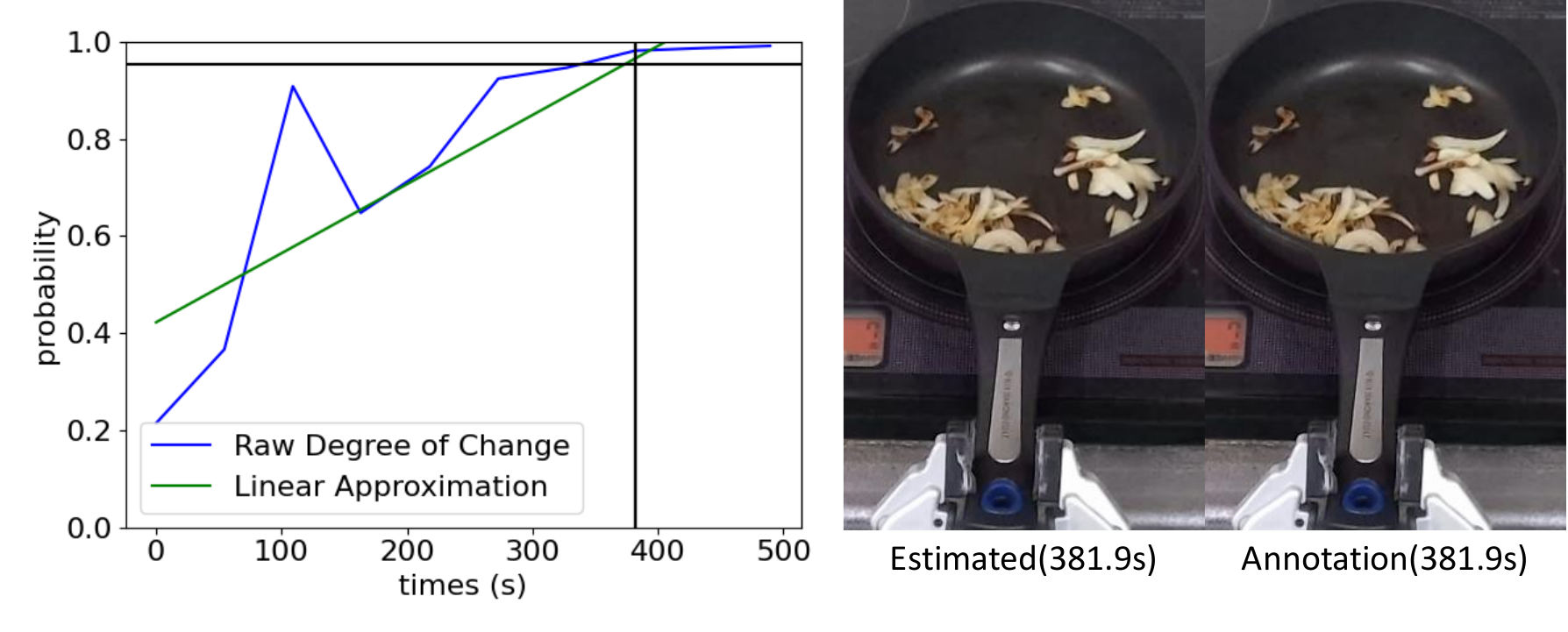}
      \centering different-power-(b)-entire
    \end{minipage}
    %% \hspace{0.01\columnwidth}
    \begin{minipage}{0.48\columnwidth}
      \includegraphics[width=\columnwidth]{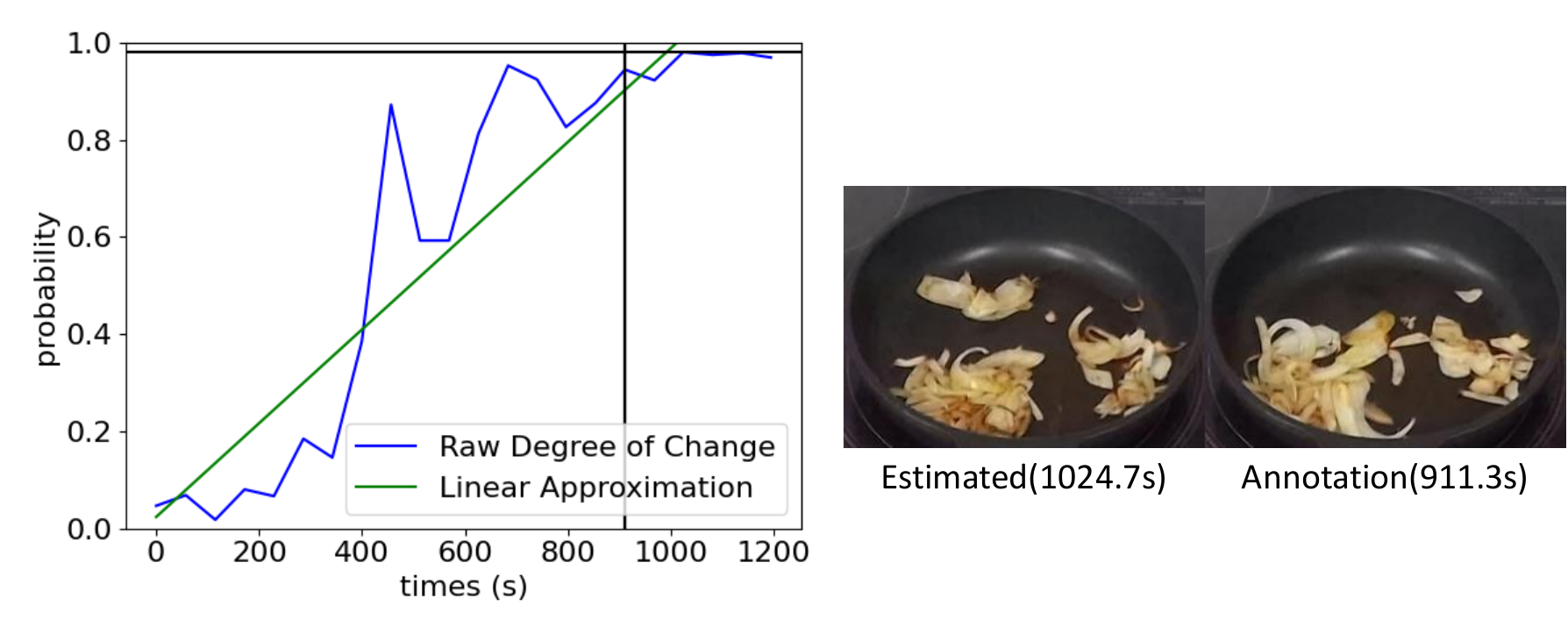}
      \centering same-power-(b)-contents
    \end{minipage}
    %% \hspace{0.01\columnwidth}
    \begin{minipage}{0.48\columnwidth}
      \includegraphics[width=\columnwidth]{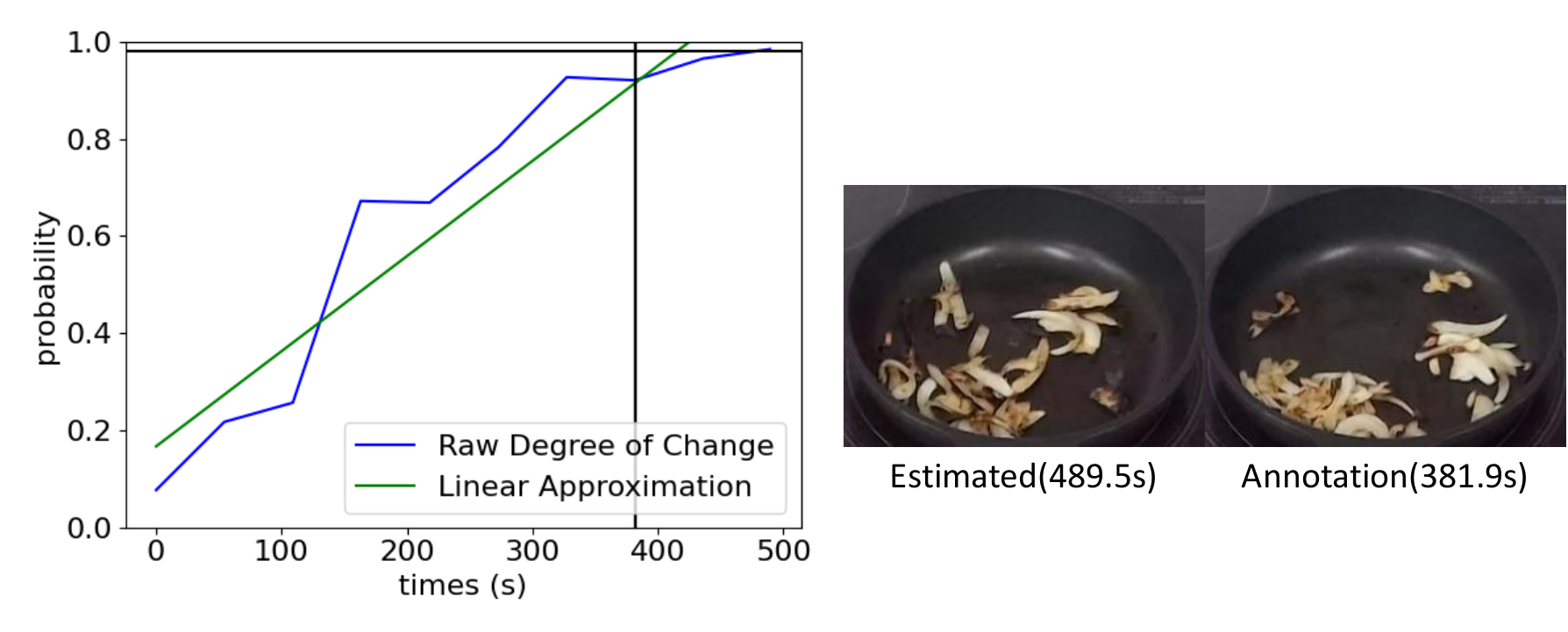}
      \centering different-power-(b)-contents
    \end{minipage}
    %% \hspace{0.01\columnwidth}
  \end{center}
  \vspace{-6mm}
  \caption{Plots of state change recognition results for unknown Maillard reaction data and comparison of images at the estimated time and at the time annotated by human.}
  \label{figure:result_image_maillard}
  \vspace{-14mm}
\end{figure}

%% \subsection{Discussion of Experiment Results}
\subsubsection{Discussion of Experiment Results}
%% 考察
%% まず，視覚-言語モデルのプロンプトについて，4種類の言語記述を用意して比較を行った．気化のデータの鍋の中身に注視した条件では(a)の単純な言語記述が最適なプロンプトだと選択されたが，それ以外の条件では(b)の食材名が最後にくる変化の様子についての記述が含まれるプロンプトで性能が最もよくなるという結果になった．概ね仮設通りに，単純な記述よりも変化についての記述が含まれる方が性能がよくなるという結果であった．また，食材名の位置については，食材名が最後にくる方が言語記述の単語数が短くなり視覚-言語モデルにとって解釈しやすいプロンプトになっている可能性が考えられる．

First, we compared four types of language descriptions for the prompts of the vision-language model. In the condition in which the focus was on the contents of the pot of vaporization data, prompt (a), the simple description, was selected as the best prompt, but in all other conditions, prompt (b), which included the description of the change caused by the state change with the ingredient word at the end, performed the best. As expected, the performance was better when the description of the change was included than when only the simple description was included. In addition, the position of the ingredient word at the end of the prompt may be easier to interpret for the vision-language model because the number of words in the language description is shorter.

%% 次に，画像の注視領域については，状態変化する対象物のみが写っている鍋やフライパンの中身を注視した条件の方が鍋全体を注視する場合よりも性能が良くなるという仮設を立てた．しかし，実験結果を見てみると，鍋全体を注視した場合の方が線形近似の傾きも大きく，認識器としての性能評価も概ね結果がよくなっている．
%% これは，視覚-言語モデルの事前学習に使われたデータの中には鍋の中身のみを写した画像よりも鍋全体を写した画像の方が多く含まれてたという可能性や，鍋の中身のみの画像よりも鍋全体が写っている画像の方が食材を調理している画像であるということが解釈しやすい画像になっている可能性などが考えられる．

%% この理由としては，視覚-言語モデルの学習に使われたデータの中には鍋の中身のみを写した画像よりも鍋全体を写した画像の方が多く含まれてた可能性や，鍋の中身のみの画像よりも鍋全体が写っている画像の方が食材を調理している画像であるということが解釈しやすい画像になっている可能性などが考えられる．

Next, for the gazing region of the image, we hypothesized that the performance would be better in the condition in which the contents of the pot or pan were gazed at than in the condition in which the entire pot or pan was gazed at, since only the state-changing object was captured in the image. However, the experimental results show that the slope of the linear approximation is larger when the entire pot or pan is gazed at, and the performance evaluation as a recognizer is also generally better.
The reasons for this may include the possibility that the data used to train the vision-language model included more images of the entire pot than images of only the contents of the pot, or that images of the entire pot are easier to interpret as images of food being cooked than images of only the pot's contents.

\section{CONCLUSIONS}

%% 概ねこのような感じでいいから，提案したこと，検証したこと，それで分かったこと．bが良い，全体が良い，卵は難しい．次に今後の展望．

%% 本研究では，調理ロボットの加熱調理における多様で特殊な状態変化の認識という課題に着目し，オープンボキャブラリな物体識別を行うことが可能な視覚-言語モデルを時系列利用することで，自然言語をプロンプトとした統一的な視覚状態変化認識法を提案した．その際に重要となるプロンプトについて，4種類の言語記述を考えて実ロボットデータによる比較を行い，状態変化により起こる変化に関する記述も追加した食材単語で終わる形の言語記述がプロンプトに適していることを確認した．

%% 加熱調理における代表的な食材の状態変化として，物理的な変化である気化と融解，化学的な変化であるタンパク質の熱変性とメイラード反応の4つの状態変化を考えた．それぞれの状態変化の調理中のデータを実ロボットにより収集し，提案手法の有効性を検証した．気化と融解，メイラード反応については状態変化認識を行えることを確認したが，タンパク質の熱変性については本研究の手法のみでは認識が難しく，他のプロンプトの利用や時系列処理方法の改善等も含めたよりロバストな認識器の設計法が必要であると考えられる．また，画像の注視領域については，鍋やフライパンの中身のみを注視するよりも鍋やフライパン全体を注視する方が概ね良い認識結果になることもわかった．

%% 今後は今回得られた知見をもとに，よりロバストな状態変化認識器の設計法を明らかにするとともに，レシピから調理実行計画法等とともに調理ロボットシステムとして統合していく．

In this study, we focused on the problem of recognizing various special state changes in the heating cooking process of cooking robots, and proposed a unified visual state change recognition method using natural language as the prompt by time-series use of the vision-language model that can perform open vocabulary object classification. We compared four types of language descriptions for the prompts, which are important in this process, using real robot data, and confirmed that language descriptions in the form of ingredient word ending, including descriptions of changes caused by the state changes, are suitable for the prompts.

We considered four typical state changes of foodstuffs during cooking: vaporization and melting, which are physical changes; thermal denaturation of proteins and maillard reactions, which are chemical changes. We collected data on each of these state changes during cooking using an actual robot, and verified the effectiveness of the proposed method. We confirmed that the proposed method can recognize vaporization, melting, and Maillard reaction, but thermal denaturation of protein was difficult to recognize only with the proposed method. It is considered necessary to design a more robust recognizer that includes a method of searching for more suitable prompts and improvement of time series processing methods. In order to recognize more types of state changes, it may be necessary to integrate them into a multimodal recognition method. In addition, it was also found that gazing at the entire pot or pan generally produced better recognition results than gazing only at the contents of the pot or pan.

Based on the findings obtained in this study, we will clarify a design method for a more robust state change recognizer and integrate it into a cooking robot system together with a cooking execution planning method based on recipes.

\bibliographystyle{unsrt}
\bibliography{main}

\end{document}